\documentclass[10pt]{article}

\def \QED{\mbox{\rule[0pt]{0.7em}{0.7em}}}

\newtheorem{lemma}{Lemma}
\newtheorem{theorem}{Theorem}

\newtheorem{remark}{Remark}

\usepackage{cite}
\usepackage{easybmat}
\usepackage{amsmath}
\usepackage{amssymb}
\usepackage{graphicx}
\usepackage{color,soul}
\def \QED{\mbox{\rule[0pt]{0.7em}{0.7em}}}

\title{\bf Distributed Value Function Approximation for Collaborative Multi-Agent Reinforcement Learning}

\author{
Milo\v{s} S. Stankovi\'{c}\thanks{e-mail: milstank@gmail.com}, Marko Beko and Srdjan S. Stankovi\'{c}}

\begin{document}

\maketitle

\begin{abstract}                % Abstract of not more than 250 words.
In this paper we propose several novel distributed gradient-based temporal difference algorithms for multi-agent off-policy learning of linear approximation of the value function in Markov decision processes with strict information structure constraints, limiting inter-agent communications to small neighborhoods. The algorithms are composed of: 1) local parameter updates based on single-agent off-policy gradient temporal difference learning algorithms, including eligibility traces with state dependent parameters, and 2) linear stochastic time varying consensus schemes, represented by directed graphs. The proposed algorithms differ by their form, definition of eligibility traces, selection of time scales and the way of incorporating consensus iterations. The main contribution of the paper is a convergence analysis based on the general properties of the underlying Feller-Markov processes and the stochastic time varying consensus model. We prove, under general assumptions, that the parameter estimates generated by all the proposed algorithms weakly converge to the corresponding ordinary differential equations (ODE) with precisely defined invariant sets. It is demonstrated how the adopted methodology can be applied to temporal-difference algorithms under weaker information structure constraints. The variance reduction effect of the proposed algorithms is demonstrated by formulating and analyzing an asymptotic stochastic differential equation. Specific guidelines for communication network design are provided. The algorithms' superior properties are illustrated by characteristic simulation results.
\end{abstract}

%\begin{keyword}
%Reinforcement learning, Distributed consensus, Multi-agent systems, Value function, Eligibility Traces, Convergence analysis, %Linear approximation, Learning algorithms, Off-policy learning, Weak convergence, Collaborative networks.
%\end{keyword}

%===============================================================================

\section{Introduction}

Interest in \emph{decentralised multi-agent algorithms} for automatic decision making in uncertain and dynamically changing environments has recently been dramatically increased, mainly due to the fundamental role of these algorithms in design and operation of the cutting edge technologies and concepts such as \emph{Cyber-Physical Systems} \emph{(CPS), Internet of Things (IoT), Swarm Robotics, Smart Grids, Smart Mobile Networking, Industry 4.0}, etc. \emph{Distributed estimation, optimization and adaptation} methods play an essential role in the development of these algorithms; a large class of them is based on dynamic collaboration, often aimed at achieving \emph{consensus} on certain variables (e.g. \cite{stankovic_calib_tcns,tsi,tba,ky1,no,bfh,timeautom,neol,ts,weak,robust_cons_stankovic_scl} and references therein). The main underlying idea is to use an inter-agent communication network (typically Wireless Sensor Network - WSN) to achieve global consensus in a completely \emph{decentralised and distributed} way. %without presence of any type of fusion center.

\emph{Reinforcement learning} (RL) is a powerful methodology for decision making in uncertain environments, which typically uses \emph{Markov Decision Process} (MDP) modeling, providing efficient approximate solutions to complex optimization problems involving dynamic programming \cite{sutton1998reinforcement,tsitsiklis1996neuro}. One of the most important tools generated within the RL field is \emph{temporal-difference} (TD) learning, typically used to learn approximations of the \emph{value function} of a given MDP \cite{sutton1998reinforcement,tvr}. This problem is especially acute in complex systems with very large state space and presence of uncertainty. It is frequently desirable to evaluate a given \emph{target policy} by implementing different \emph{behavior policies} (off-policy learning, e.g. \cite{precup2001off}). In \cite{sutton2009fast,geist_survey,maei2011,dann2014, yu_off_policy_2017,dai_nonlinear_off_policy_2017} several \emph{fast gradient-based algorithms} for TD learning have been proposed, successfully handling most practical aspects.

\emph{Distributed multi-agent RL methods} have received recently a lot of attention due to their high potential for solving essential problems within complex, intelligent and networked systems belonging to CPS, IoT, Swarm Robotics and the other mentioned emerged areas (see, e.g. \cite{busoniu2008comprehensive,gupta_survey_2017,OroojlooyJadid_review_2019} and numerous references therein). The problem of \emph{distributed multi-agent value function approximation} has attracted great attention either \emph{per se}, e.g.  \cite{swmr,mathkar2013distributed,valcarcel2015distributed,naira_cdc_2018,cassano_sayed_linear_rate_ECC_2019,stankovic_ACC_2016,doan_finite_time_2019,leehu2020,dwywj2020}  or within the \emph{actor/critic} algorithms, as their \emph{critic} part, e.g. \cite{pp,zhang_zavlanos_2019,basar2019,zhang2018fully}. Typically, a specific distributed setup is adopted in which it is assumed that each agent can access (observe transitions) of a given MDP independently, without mutual interactions with other agents through the MDP environment. We have adopted in this paper the line of thought which has connections to the following contributions related to the distributed multi-agent RL relying on consensus-like collaborations between the agents: \cite{swmr}, from the point of view of information structure constraints (ISC); \cite{valcarcel2015distributed}, where the mean square convergence of a distributed gradient based algorithm without eligibility traces has been proved, assuming independent sampling; \cite{mathkar2013distributed}, where a proof of almost sure convergence of a distributed on-policy TD(0) algorithm is provided for gossip-like communications; \cite{leehu2020,dwywj2020}, where weaker ISC allow a continuous insight into the states and actions of all MDP's in the system. The last assumption related to the weak ISC applies to the majority of the available distributed actor/critic algorithms  \cite{pp,zhang_zavlanos_2019,basar2019,zhang2018fully}, as well as the approach to distributed RL proposed in \cite{kar2013learning} where estimates of all the possible state-action pairs (which is typically very large) of the so-called $Q$ function, are maintained, not involving any parametric approximation which is essential for this paper.

The main general motivation for this paper has been the desire to provide new tools, with strictly provable properties, for efficient collaborative exploration of large state spaces and for variance reduction under strict ISC, as well as for computation parallelization. We propose several new algorithms for \textit{distributed} \textit{multi-agent off-policy gradient temporal difference learning} \emph{of linear approximation of the value function} in MDPs, starting from the single-agent off-policy gradient-based algorithms proposed in \cite{maei2011,sutton2009fast,maeisutton,geist_survey,yu_off_policy_2017,dann2014}, and using linear dynamic \emph{consensus iterations} based on local communications according to strict ISC. The algorithms differ among themselves by the definition of \emph{(state dependent) eligibility traces}, by the way in which the consensus scheme is applied, as well as by the way in which the time scales are introduced \cite{maei2011,sutton2009fast,maeisutton,yu_off_policy_2017,dann2014}. Only one of the algorithms that we propose is a generalization to the one presented in \cite{valcarcel2015distributed}; the remaining ones can be considered as new. Assuming general stochastic time-varying dynamic consensus scheme and nonrestrictive assumptions concerning the MDP properties, a rigorous \emph{proof of the weak convergence of parameter estimates to consensus} is provided for all the proposed algorithms, based on appropriately defined ordinary differential equations (ODE's) with specified \emph{limit sets}  \cite{ky1,ky,yu_off_policy_2017,weak}; this proof represents the central point of the paper. The proof is based on general properties of the Feller-Markov chains \cite{yu_off_policy_2017} and the properties of distributed stochastic approximation \cite{ky1,weak,ky}; notice that the algorithms discussed in \cite{sutton2009fast,naira_cdc_2018,valcarcel2015distributed,stankovic_ACC_2016} are based on unrealistic data independence assumptions. The \emph{weak convergence methodology} has been adopted, having in mind its intuitive appeal closely connected to practical reasoning and the fact that the imposed restrictions are by far weaker than in the case of alternative methodologies \cite{ky,bor,bm,yu_off_policy_2017}. It will be shown that the proposed methodology of algorithm design and convergence analysis can be extended to the case of weaker ISC, adopted in the algorithms from, \emph{e.g.}, \cite{leehu2020,dwywj2020,pp,zhang_zavlanos_2019,basar2019,zhang2018fully}. The effect of variance reduction introduced by the proposed algorithms is verified by an analysis based on a construction of a stochastic differential equation (SDE) which models the asymptotic behavior of the estimates. Specific guidelines are given on how to design the communication network in order to ensure the desired sets of convergence points and fast convergence rate. Finally, selected simulation results illustrate the main concepts and properties of the algorithms, providing a comparison which demonstrates superiority of the proposed schemes compared to the existing ones.

The paper is organized as follows. In Section~\ref{sec:problem_formulation} we formulate the problem and define the algorithms. The first part of Section~\ref{sec:conv} is devoted to preliminary results, including some basic properties of the Feller-Markov state-trace processes and of the incorporated consensus scheme. In the second part of Section~\ref{sec:conv} a proof of weak convergence to consensus is presented for all the proposed algorithms. Section~\ref{sec_discussion} is devoted to a discussion on several important issues, such as a possibility to introduce constraints on the parameter vector, the overall impact of consensus and the application of the algorithm in the case of weaker ISC, the communication network design, and the variance reduction effect.
Finally, in Section~\ref{sec:sim} the results of simulations are shown.

\section{Distributed Gradient Based Temporal Difference Algorithms} \label{sec:problem_formulation}
\subsection{Problem Formulation and Definition of the Algorithms}

Consider $N$ \emph{autonomous agents}, each acting on a separate Markov Decision Process (MDP), denoted as MDP$^{(i)}$, $i=1, \ldots, N$, characterized by the quadruplets $\mathcal{Q}_{i}=\{\mathcal{S},$ $\mathcal{A}, $ $p(s'|s,a),R_{i}(s,a,s')\}$, where $\mathcal{S}=\{s_{1}, \ldots, s_{M} \}$ is a finite set of states,  $\mathcal{A}$ is a finite set of actions, $p(s'|s,a)$ is defining probabilities of moving from $s \in \mathcal{S}$ to $s' \in \mathcal{S}$ by applying action $a \in \mathcal{A}$, and $R_{i}(s,a,s')$ are random rewards distributed according to $q(\cdot|s',a,s)$; let MDP$^{(0)}$ represent a fictitious reference MDP characterized by $\mathcal{Q}_{0}$. Each MDP$^{(i)}$, $i=0,1, \ldots, N$, applies a fixed stationary \emph{behavior policy} $\pi^{(i)}(a|s)$ (probability of taking action $a$ at state $s$), implying that the state processes $\{S_{i}(n) \}$ and the state-action processes $\{S_{i}(n), A_{i}(n)\}$, where $n \geq 1$ is an integer denoting transition time, represent time homogenous Markov chains. The goal of the agents is to learn the \emph{state value function} for a given \emph{target policy}  $\pi^{(0)}=\pi$ formally corresponding to MDP$^{(0)}$, using the information of state transitions and rewards in MDP$^{(i)}$, $i=1, \ldots, N$. Therefore, we are dealing with a \emph{cooperative} \emph{off-policy} \emph{reinforcement learning} problem.
\par
Let $P^{(i)}$ denote the transition matrices of the Markov chains $\{S_{i}(n) \}$, with $P^{(i)}_{ss'}$ being the probabilities of transitions from state $s\in \mathcal{S}$ to $s'\in \mathcal{S}$,  $i=0, \ldots, N$.
The desired \emph{state value function} is defined using \emph{discount factors} $\gamma(s) \in [0,1]$, $s \in \mathcal{S}$ \cite{sutton1998reinforcement, yumasu}. If the expected discounted total reward is denoted as $v_{\pi}(s)$, $s \in \mathcal{S}$,  the $M$-vector $v_{\pi}=[v_{\pi}(s_1) \cdots v_{\pi}(s_M)]^{T}$, defining the value function for all $s \in \mathcal{S}$, uniquely satisfies the \emph{Bellman equation}
\begin{equation} \label{Bel0}
 v_{\pi}=r_{\pi}+P\Gamma v_{\pi},
\end{equation}
where $r_{\pi}=[r_{\pi}(s_1) \cdots r_{\pi}(s_M)]^{T} $,  $r_{\pi}(s)$ representing the one-stage expected rewards at each state $s \in \mathcal{S}$ under policy $\pi$, $P=P^{(0)}$ and $\Gamma$ denotes the $M \times M$ diagonal matrix with $\gamma(s)$, $s \in \mathcal{S}$, as diagonal entries. Besides (\ref{Bel0}), $v_{\pi}$ also satisfies a family of \emph{generalized Bellman equations}, $v_{\pi}=T^{(\lambda)} v_{\pi}$, where $T^{(\lambda)}$ is the \emph{generalized Bellman operator} $T^{(\lambda)} v=r_{\pi}^{(\lambda)}+P^{(\lambda)}v$, $\forall v \in \mathcal{R}^{M}$, for a given vector $r_{\pi}^{(\lambda)}$ and a substochastic matrix $P^{(\lambda)}$, where $\lambda \in [0,1]$ are the so-called $\lambda$-parameters \cite{yu_off_policy_2017,yumasu}. Analogously, the affine Bellman operators for MDP$^{(i)}$, $i=1, \ldots, N$, can be defined as $T^{(\lambda_{i})}$, with vector $r_{\pi}^{(\lambda_{i})}$ and a substochastic matrix $P^{(\lambda_{i})}$.
Introduce the local \emph{importance sampling ratios} $\rho_{i}(s,s')=P_{ss'}/P^{(i)}_{ss'}$ for $s,s' \in \mathcal{S}$ (with $0/0=0$). The following assumption ensures well defined value function and importance ratios \cite{sutton1998reinforcement,stankovic_ACC_2016}:

(A1) (\emph{Assumptions on target and behavior policies})
\newline
a) $P$ is such that $I-P\Gamma$ is nonsingular;
\newline
b) $P^{(i)}$ are irreducible and such that for all $s,s' \in \mathcal{S}$, $P^{(i)}_{ss'}=0$ $\Rightarrow$ $P_{ss'}=0$, $i=1, \ldots, N$.

Let $\phi: \mathcal{S} \rightarrow \mathcal{R}^{p}$ be a function that maps each state to a $p$-dimensional feature vector $\phi$; let the subspace spanned by these vectors be $\mathcal{L}_{\phi}$. Our goal is to find $v =[v(s_1) \cdots v(s_M)]^T $ $ \in \mathcal{L}_{\phi}$ that satisfies $v \approx T^{(\lambda)} v$. Introduce $v_{\theta}=\Phi \theta$, where $\Phi$ is an $M \times p$ matrix composed of $p$-vectors $\phi(s)$ as row vectors and $\theta \in \mathcal{R}^{p}$ is a parameter vector.  %that the algorithm should learn the vector $\theta$.
\par
%In order to construct a distributed algorithm for collaborative estimation of the parameter vector $\theta$
%by using observations from MDP$^{(i)}$, $i=1, \ldots, N$, we
Introduce the global parameter vector $\Theta=[\theta_{1}^{T} \cdots \theta_{N}^{T}]^{T}$  and define the following \emph{constrained optimization problem}
\begin{eqnarray} \label{j}
& \mathrm{Minimize} \; J(\Theta)= \sum_{i=1}^{N} q_{i} J_{i}(\theta_{i}) \\
& {\rm Subject} \; {\rm to} \;\; \theta_{1}= \cdots = \theta_{N}=\theta, \nonumber
\end{eqnarray}
where $J_{i}(\theta_{i})=  \| \Pi_{\xi_{i}}\{T^{(\lambda_{i})} v_{\theta_{i}}-v_{\theta_{i}}\} \|^{2}_{\xi_{i}}$ are the \emph{local objective functions}, $q_{i} > 0$ \emph{ a priori }defined weighting coefficients,
 $\lambda_{i}$ the local $\lambda$-parameters and  $ \Pi_{\xi_{i}}\{ \cdot \}$  the projection onto the subspace $\mathcal{L}_{\phi}$ w.r.t. the weighted Euclidean norm $\|v\|^{2}_{\xi_{i}}= \sum_{s \in \mathcal{S}} \xi_{i;s} v(s)^{2}$ for a positive $M$-dimensional vector $\xi_{i}$ with components $\xi_{i;s}$, $s=s_1, \ldots, s_M$ (see \cite{stankovic_ACC_2016,yu_off_policy_2017}). In accordance with \cite{yu_off_policy_2017,yumasu}, we take $\xi_{i}$ to be the \emph{invariant probability distribution} for the local Markov chain $\{S_{i}(n)\}$, with the transition matrix $P^{(i)}$ induced by $\pi^{(i)}$, satisfying $\xi_{i}^{T} P^{(i)}= \xi_{i}^{T}$, $i=1, \ldots, N$. It follows that
\begin{equation} \label{grad1}
\nabla J(\theta)=\sum_{i=1}^{N} q_{i} (\Phi^{T} \Xi_{i} (P^{(\lambda_{i})}-I) \Phi)^{T} w_{i}(\theta),
\end{equation}
where $\nabla J(\theta)= \nabla J(\Theta)|_{\theta_{1}= \cdots = \theta_{N} = \theta}$, $\Xi_{i}$ is the $M \times M$ diagonal matrix with the components of $\xi_{i}$ on the diagonal, and $w_{i}(\theta)$ represents the unique solution (in $w_{i}$) of the equation
%\begin{equation} \label{weq}
$\Phi w_{i}=\Pi_{\xi_{i}} \{ T^{(\lambda_{i})} v_{\theta}-v_{\theta} \}$,
%\end{equation}
assuming that $w_{i} \in {\rm span} \{\phi(\mathcal{S})\}$; it is possible to show that this equation is equivalent to $\Phi^{T}\Xi_{i} \Phi w_{i}=\Phi^{T}\Xi_{i} (T^{(\lambda_{i})} v_{\theta}-v_{\theta})$ \cite{yu_off_policy_2017}.
\par
Alternatively, one can reformulate (\ref{grad1}) in the following way:
\begin{align} \label{grad2}
\nabla J(\theta)=\sum_{i=1}^{N} q_{i}[&-\Phi^{T} \Xi_{i} (T^{(\lambda_{i})}v_{\theta}-v_{\theta}) + (\Phi^{T} \Xi_{i} P^{(\lambda_{i})}\Phi)^{T} w_{i}(\theta)].
\end{align}
%We define "local" GTD (gradient temporal-difference) updates of the local parameter vectors using samples from the expressions for the local gradients (\ref{grad1}) and (\ref{grad2}).
\par
Let $\rho_{i}(n)=\rho_{i}(S_{i}(n),S_{i}(n+1))$ and $\gamma_{i}(n)=\gamma(S_{i}(n))$;
\begin{align}
\delta_{i}(v_{\theta};n)=\rho_{i}(n)(&R_i(n+1) +\gamma_i(n+1)v_{\theta}(S_{i}(n+1))-v_{\theta}(S_{i}(n))) 
\end{align} represents the local \emph{temporal-difference term} \cite{yu_off_policy_2017,yumasu}.

We propose below several algorithms composed of \emph{two main parts}: 1) \emph{local parameter updates}, based on the \emph{gradient descent} methodology developed for single-agent case, using local state transition and reward observations from MDPs, and 2) \emph{convexification} of current parameter estimates based on inter-agent communications.

We first propose two algorithms, which differ in the first part. The first one is derived from (\ref{grad1}) and denoted as D1-GTD2($\lambda$) (according to the GTD2 algorithm  proposed in \cite{sutton2009fast})
\begin{align}
\theta'_{i}(n)=& \theta_{i}(n)+ \alpha_i(n) q_{i} \rho_{i}(n)(\phi(S_{i}(n)) -\gamma_{i}(n+1) \phi(S_{i}(n+1))) e_{i}(n)^{T}w_{i}(n) \label{a}  \\
w'_{i}(n)=& w_{i}(n)+ \beta_i(n)(e_{i}(n) \delta_{i}(v_{\theta_{i}(n)};n) -\phi(S_{i}(n)) \phi(S_{i}(n))^{T} w_{i}(n)), \label{w}
\end{align}
and the second one derived from (\ref{grad2}), denoted as D1-TDC($\lambda$) (according to the TDC algorithm from \cite{sutton2009fast})
\begin{align} 
\theta'_{i}(n)=&\theta_{i}(n)+ \alpha_i(n)q_{i} [e_{i}(n) \delta_{i}(v_{\theta_{i}(n)};n) -\rho_{i}(n) \label{TDC} \times  \\ &(1-\lambda_{i}(n+1)) \gamma(n+1) \phi(S_{i}(n+1)) e_{i}(n)^{T} w_{i}(n)],  \nonumber
\end{align}
with the same relation for $w_{i}'(n)$ given by (\ref{w});
in (\ref{a}), (\ref{w}) and (\ref{TDC}),  $v_{\theta_i(n)}=v_{\theta}(S_{i}(n))|_{\theta=\theta_{i}(n)}=v_{\theta_{i}(n)}(S_{i}(n))=
\phi(S_{i}(n))^{T} \theta_{i}(n)$, and $e_{i}(n)$ is the \emph{eligibility trace} vector generated by
\begin{equation} \label{elig_rec}
e_{i}(n)=\lambda_i(n)\gamma_i(n)\rho_{i}(n-1)e_{i}(n-1)+\phi(S_{i}(n)).
\end{equation}
%with $\lambda_i(n)$ being the state dependent $\lambda$-parameters.
The initial values $\theta_{i}(0)$ are chosen arbitrarily; however, $w_{i}(0)$, as well as $e_{i}(0)$, have to satisfy $w_{i}(0)$, $e_{i}(0)$ $\in {\rm span} \{ \phi(\mathcal{S}) \}$  \cite{yu_off_policy_2017}.
%The second, denoted as ALGb, is derived after adding importance sampling and eligibility traces to the algorithm TDC from [.]. Its local updates are generated by:
%\par
%\begin{eqnarray} \label{b}
%&\theta'_{i}(n)=\theta_{i}(n)+ \alpha_{i}(n)q_{i} [e_{i}(n) \delta_{i}(v_{\theta_{i}(n)};n)- & \\
%& \rho_{i}(n)(1-\lambda_{i}(n+1) \gamma(n+1) \phi(S_{n+1}) e_{i}(n)^{T} w_{i}(n)], & \nonumber
%\end{eqnarray}
%with the recursion for $w_{i}(n)$ identical to the one in (\ref{a}), together with the corresponding initial conditions.
Sequences $\{\alpha_i(n)\}$ and $\{\beta_i(n)\}$ are positive step size sequences, which can be either of the same order of magnitude (single-time-scale) or satisfying $\alpha_i(n) << \beta_i(n)$ (two-time-scale), see \cite{yu_off_policy_2017}.
%In both formulae, $\{\alpha_{i}(n)\}$ and $\{\beta_{i}(n)\}$ are positive step sizes, which can be either equal order of magnitude (single time-scale), or satisfying $\alpha_{i}(n) << \beta_{i}(n)$ (two time-scales)  \cite{yu_off_policy_2017,sutton2009fast}.
%Following the adopted methodology of constructing \emph{consensus-based distributed parameter and state estimation algorithms} (in accordance with the constraint in (\ref{j})),

The second part of the algorithms is given, for both D1-GTD2($\lambda$) and D1-TDC($\lambda$),  by
\begin{equation} \label{partial}
\theta_{i}(n+1)= \sum_{j=1}^{N} a_{ij}(n) \theta'_{j}(n),  \;\;\;
 w_{i}(n+1)=  w'_{i}(n).
\end{equation}
If we apply the consensus convexifications also to $w_{i}(n)$, instead of (\ref{partial}), we have
\begin{equation} \label{full}
\theta_{i}(n+1)= \sum_{j=1}^{N} a_{ij}(n) \theta'_{j}(n);  \;\;
w_{i}(n+1)= \sum_{j=1}^{N} a_{ij}(n) w'_{j}(n),
\end{equation}
and we denote the corresponding algorithms as D2-GTD2($\lambda$) and D2-TDC($\lambda$).
In (\ref{partial}) and (\ref{full}), $a_{ij}(n) \geq 0$ are random variables, elements of a time-varying random matrix $A(n)=[a_{ij}(n)]$ \cite{weak,stankovic_ACC_2016}.
If one adopts that the  agents are connected by communication links in accordance with a directed graph $\mathcal{G}=(\mathcal{N},\mathcal{E})$, where $\mathcal{N}$ is the set of nodes and $\mathcal{E}$ the set of arcs, then matrix $A(n)$, has zeros at the same places as the graph adjacency matrix $A_{\mathcal{G}}(n)=A_{\mathcal{G}}$, and is \emph{row-stochastic,} \emph{i.e.} $\sum_{j=1}^{N} a_{ij}(n)=1,$ $i=1, \ldots, N$, $\forall n \geq 0$.

\section{Convergence Analysis} \label{sec:conv}
\subsection{Preliminaries}
\subsubsection{Properties of the State-Trace Processes} \label{subsec:prelims}
\par
The state-trace processes  $\{S_{i}(n), e_{i}(n)\}$ are  Markov chains with the weak Feller property (see \cite{yu_off_policy_2017,yumasu} for details). In order to formulate candidates for the asymptotic mean ODE's that should be attached to the above algorithms, define $Z_{i}(n)=$ $(S_{i}(n), e_{i}(n),$ $ S_{i}(n+1))$ $ \subset \mathcal{Z}_{i}$.
According to (\ref{a}), for D1-GTD2($\lambda$) and D2-GTD2($\lambda$), after denoting $z=(s,e,s')$, we introduce functions
\begin{equation} \label{gia}
g_{i}(\theta, w, z)=\rho_{i}(s,s')(\phi(s)-\gamma(s')\phi(s'))e^{T} w
\end{equation}
 and
\begin{equation} \label{ki}
 k_{i}(\theta, w, z)=e \bar{\delta}_{i}(s,s',v_{\theta})-\phi(s) \phi(s)^{T} w,
\end{equation}
 where $ \bar{\delta}_{i}(s,s',v_{\theta})=
\rho_{i}(s,s')(r_{i}(s,s')+\gamma(s')v_{\theta}(s')-v_{\theta}(s))$ and $r_{i}(s,s')$ is the one-step expected reward following policy $\pi^{(i)}$ when transitioning from  $s$ to $s'$. Notice that $\delta_{i}(v_{\theta_{i}(n)};n)$ and $\bar{\delta}_{i}(S_i(n), S_i(n+1), v_{\theta_{i}(n)})$ differ by the zero-mean \emph{noise term} $e_{i}(n) \omega_{i}(n+1)$, where
\begin{equation}
\omega_{i}(n+1)=\rho_{i}(n)(R_{i}(n+1)-r_{i}(S_{i}(n),S_{i}(n+1))).
\end{equation}
We have further
\begin{align}
& \bar{g}_{i}(\theta, w)=  (\Phi^{T} \Xi_{i}(I-P^{(\lambda_{i})}) \Phi)^{T} w,&  \label{barg} \\
&\bar{k}_{i}(\theta, w)=\Phi^{T} \Xi_{i} (T^{(\lambda_{i})}v_{\theta}-v_{\theta})- \Phi^{T} \Xi_{i} \Phi w. & \label{bark}
\end{align}
As for any given $\theta_{i}$ there is a unique solution $w=w_{\theta_i}=\bar{w}_{i}(\theta_{i})$ to the linear equation $\bar{k}_{i}(\theta_{i}, w)=0$, $w \in {\rm span} \{\phi(\mathcal{S}) \}$,  we obtain
%\begin{equation} \label{bargi}
%\bar{g}_{i}(\theta_{i}, w_{i}(\theta_{i}))=(\Phi^{T} \Xi_{i}(I-P^{(\lambda_{i})}) \Phi)^{T} w_{i}(\theta_{i}).
%\end{equation}
 that  $\bar{g}_{i}(\theta_{i}, \bar{w}_{i}(\theta_{i}))=- \nabla J_{i}(\theta_{i})$ (see (\ref{grad2})).
In the case of D1-TDC($\lambda$) and D2-TDC($\lambda$), we have
\begin{align} \label{gib}
g_{i}(\theta, w, z)=&e\bar{\delta}_i(s,s',v_{\theta})-\rho_{i}(s,s') (1-\lambda_{i}(s')) \gamma(s') \phi(s')e^{T}w,
\end{align}
together with the corresponding mean values. %Convergence proofs given below will be based on (\ref{barg}), (\ref{bark}) and (\ref{gib}).
%Following \cite{yu_off_policy_2017}, it is straightforward to show that in this case we obtain the same result as in (\ref{barg}), with $w_{i}=w_{i}(\theta_{i})$.
%\begin{align} \label{bargb}
%\bar{g}_{i}(\theta_{i}, w_{i}(\theta_{i}))&=\Phi^{T} \Xi_{i} (T^{(\lambda_{i})}v_{\theta_{i}}-v_{\theta_{i}}) \nonumber \\ &\quad-
%(\Phi^{T} \Xi_{i}P^{(\lambda_{i})} \Phi)^{T} w_{i}(\theta_{i}) \nonumber \\ &=- \nabla J_{i}(\theta_{i}).
%\end{align}
\par
The following result is fundamental for our analysis:
\begin{lemma}[\cite{yu_off_policy_2017}] \label{lemma:1}
Under (A1), the following holds for each $\theta_{i}$ and $w_{i}$ and each compact set $D_{i} \subset \mathcal{Z}_{i}$:

a) $\lim_{m, n \to \infty} \frac{1}{m} \sum_{s=n}^{n+m-1} E_{n} \{k_{i}(\theta_{i}, w_{i}, Z_{i}(s))-\bar{k}_{i}(\theta_{i}, w_{i})\} I(Z_{i}(n) \in D_{i})=0$ in mean,

b) $\lim_{m, n \to \infty} \frac{1}{m} \sum_{s=n}^{n+m-1} E_{n} \{g_{i}(\theta_{i}, w_{i}, Z_{i}(s))-\bar{g}_{i}(\theta_{i}, w_i)\} I(Z_{i}(n) \in D_{i})=0$ in mean,
\newline
where $E_{n}\{ \cdot \}$ denotes the conditional expectation given $(Z_{i}(0), \ldots, Z_{i}(n), R_i(0), \ldots , R_i(n))$, $i=1, \ldots N$, and $I(\cdot)$ is the indicator function.
\end{lemma}

\subsubsection{Global Model}

Let $X(n)=[\Theta(n)^{T} \vdots W(n)^{T}]^{T}$, $\Theta(n)=[\theta_{1}(n)^{T} \cdots \theta_{N}(n)^{T}]^{T}$, $W(n)=[w_{1}(n)^{T} \cdots w_{N}(n)^{T}]^{T}$ and $X'(n)=[\Theta'(n)^{T} \vdots W'(n)^{T}]^{T}$. Then, we have 
\begin{align} \label{alg}
 X'(n)&=  X(n)+\Gamma(n) F (X(n),n), \nonumber \\  X(n+1)&= {\rm diag} \{(A(n) \otimes I_{p}), I_{Np} \}  X'(n),
 \end{align}
$X(0)=X_{0}$, where $\otimes$ denotes the Kronecker's product, while $\Gamma(n)={\rm diag} \{\alpha_{1}(n), \ldots,$ $ \alpha_{N}(n), \beta_{1}(n), \ldots, \beta_{N}(n)\} $ $\otimes I_{p},$ $F(X(n),n)=[ F^{\theta}(X(n),n)^{T} \vdots F^{w}(X(n),n)^{T}]^{T}$, $F^{\theta}(X(n),n)=[F^{\theta}_{1}(X(n),n)^{T} \cdots F^{\theta}_{N}(X(n),n)^{T}]^{T}$, $F^{w}(X(n),n)=[F^{w}_{1}(X(n),n)^{T} \cdots F^{w}_{N}(X(n),n)^{T}]^{T}$, with
$F^{\theta}_{i}(X(n),n)= q_{i} g_{i}(\theta_{i}(n), w_{i}(n), Z_{i}(n))$ and $F^{w}_{i}(X(n),n)= k_{i}(\theta_{i}(n), w_{i}(n), Z_{i}(n))+e_{i}(n) \omega_{i}(n+1)$ for the algorithms of GTD2-type, and
$F^{\theta}_{i}(X(n),n)=q_{i} g_{i}(\theta_{i}(n), w_{i}(n), Z_{i}(n))+ e_{i}(n) \omega_{i}(n+1)$ for the algorithms of TDC-type (in the latter case $g_{i}( \cdot )$ is defined by (\ref{gib}) and  $F^{w}(X(n),n)$ remains the same as in the case of GTD2-type algorithms).
For the algorithms D2-GTD2($\lambda$) and D2-TDC($\lambda$), we have a modified model (\ref{alg}), in which, instead of ${\rm diag} \{(A(n) \otimes I_{p}), I_{Np} \}$,  we have
${\rm diag} \{(A(n) \otimes I_{p}), (A(n) \otimes I_{p}) \}$. Also, we introduce $\bar{F}(X)=[\bar{F}^{\theta}(X)^{T} \vdots \bar{F}^{w}(X)^{T}]^{T},$ where $\bar{F}_{i}^{\theta}(X)=q_{i} \bar{g}_{i}(\theta,w)$ and $\bar{F}_{i}^{w}(X)=q_{i} \bar{k}_{i}(\theta,w)$, $i=1, \ldots, N$.

\subsubsection{Consensus Part}
Define $\Psi(n|k)=A(n) \cdots A(k)$  for $n \geq k$, $\Psi(n|n+1)=I_{N}$. Let $\mathcal{F}_{n}$ be an increasing sequence of $\sigma$-algebras such that $\mathcal{F}_{n}$ measures $ \{X(k), k \leq n, A(k), k < n\}$.
\par
(A2) There is a scalar $\alpha_{0} > 0$ such that $a_{ii}(n) \geq \alpha_{0}$, and, for $i \neq j$, either $a_{ij}(n)=0$ or $a_{ij}(n) \geq \alpha_{0}$.
\par
(A3) Graph $\mathcal{G}$ is strongly connected.
\par
(A4) There exist a scalar $p_{0} > 0$ and an integer $n_{0}$ such that $P_{\mathcal{F}_{n}}$\{agent $j$ communicates to agent $i$ on the interval $[n, n+n_{0}]$\} $\geq p_{0}$, for all $n$, $i=1, \ldots N$, $j \in \mathcal{N}_i$.
\par
\begin{lemma}[\cite{ky1,weak}] \label{lemma:2} Let (A2)--(A4) hold. Then $\Psi(k)=\lim_{n} \Psi(n|k)$ exists with probability 1 (w.p.1) and its rows are all equal; moreover,
$E \{ |\Psi(n|k)-\Psi(k)| \}$ and $E_{\mathcal{F}_{k}} \{ |\Psi(n|k)-\Psi(k)| \}$ $\rightarrow 0$ geometrically as $n-k \rightarrow \infty$, uniformly in $k$ (w.p.1); also, $E_{\mathcal{F}_{k}}\{ \Psi(n|k) \}$ converges to $\Psi(k)$ geometrically, uniformly in $k$, as $n \rightarrow \infty$ ($|\cdot|$ denotes the infinity norm).
\end{lemma}
\par
(A5) There is a $N \times N$  matrix $\bar{\Psi}$ such that $E\{| E_{\mathcal{F}_{k}} \{\Psi(n)\}- \bar{\Psi} | \} \rightarrow 0$ as $n-k \to \infty$, which, according to Lemma~\ref{lemma:2}, has the form
 $\bar{\Psi}=[\hat{\Psi}^{T} \cdots \hat{\Psi}^{T}]^{T}$, where $\hat{\Psi}=[\bar{\psi}_{1}  \cdots \bar{\psi}_{N}]^{T}$.
% \[\bar{\Psi}=\left[ \begin{BMAT}{ccc}{ccc}
%\bar{\psi}_{1}  & \cdots & \bar{\psi}_{N}  \\ & \cdots & \\ \bar{\psi}_{1}  & \cdots & \bar{\psi}_{N}  \end{BMAT} \right]= \left[ \begin{BMAT}{c}{ccc}
%\hat{\Psi} \\  \vdots \\ \hat{\Psi} \end{BMAT} \right], \] where $\sum_{i} \bar{\psi}_{i}=1$ .
\par
(A6) Sequence $\{A(n)\}$ is independent of the processes in MDP$^{(i)}$, $i=1, \ldots, N$.
\begin{remark}
Assumptions (A2)--(A6), formulated according to \cite{ky1}, are essentially very mild and do not impose any significant restrictions in practice. They allow different time-varying network models such as asynchronous broadcast gossip schemes including possible communication failures \cite{weak}. \end{remark}
\subsection{Convergence Proofs}
In the sequel, we pay attention to several characteristic cases. Theorems~\ref{th:1} and \ref{th:2} are related to GTD2($\lambda$) based algorithms in one-time-scale. Theorem~\ref{th:1} deals with D1-GTD2($\lambda$) (consensus only on $\theta$), whereas Theorem~\ref{th:2} deals with D2-GTD2($\lambda$) (consensus on both $\theta$ and $w$). Theorem~\ref{th:3} treats D1-GTD2($\lambda$) in two time scales. Using the preliminaries from Subsection~\ref{subsec:prelims} it is straightforward to analogously formulate convergence theorems for D1-TDC($\lambda$), D2-TDC($\lambda$) and D2-GTD2($\lambda$) (all in two time scales) and prove them using the same arguments as in the proofs of the provided theorems.
%In the sequel, we shall pay attention to several distinct cases, depending on the algorithm form.
%We shall assume that $\alpha_{i}(n)=\alpha$ and $\beta_{i}(n)=\beta$;  in the asymptotic analysis,  $\alpha, \beta  \to 0$ with $\alpha/\beta \rightarrow \varepsilon$, where $\varepsilon = 1$ for one-time-scale, and $\varepsilon = 0$
%for two-time-scales. The presented results can be extended with no difficulty to the case when $\alpha_{i}(n)$ and $\beta_{i}(n) $ are decreasing sequences tending to zero when $n \to \infty$ (see \cite{ky,yu_off_policy_2017}).
\par
(A7) Sequence $\{X(n)\}$ is tight (for definition and theoretical background see, e.g. \cite{ky}).
\begin{remark}
Assumption (A7) is frequent for weak convergence proofs in different contexts.  As stated in \cite{ky1,ky}, one can achieve, without loss of generality, that $\{X(n)\}$ is tight by adequate projection or truncation (see Subsection~\ref{subsec:constr}). In this paper, our aim is to place focus on other aspects of the convergence of the proposed algorithms.
\end{remark}
%Notice that \emph{Case a)} corresponds to ALGa (the original GTD2 algorithm from [.]) represented by the recursions (\ref{a}), (\ref{w}), with $\beta(n)=\alpha(n)=\alpha$), \emph{Case b)} to ALGa according to the algorithm GTDa %from [.], represented by the recursions (\ref{a}), (\ref{w}), with $\alpha(n)=\alpha$, $\beta(n)=\beta$, $\beta >>\alpha$), and \emph{\emph{Case c)}} to the original TDC algorithm proposed in [.] and analyzed in as GTDb [.], %represented by the recursions (\ref{b}), (\ref{w}), with $\alpha(n)=\alpha$, $\beta(n)=\beta$, $\beta >>\alpha$).
%(according to [.], the TDC algorithm has no convenient interpretation in the , according to [.])
%\subsubsection{Case A)}
%\emph{ D1-GTD2($\lambda$), $ \varepsilon=1$}

Following \cite{ky1}, let $n_{\alpha}$ be a sequence tending to $\infty$ and satisfying $\alpha^{\frac{1}{2}} n_{\alpha} \to 0$ as $\alpha \to 0$. Define
 \begin{align}
X_{0}^{\alpha}=&{\rm diag} \{\Psi(n_{\alpha}|0) \otimes I_{p}, I_{Np} \}  X_{0} \nonumber \\ &+ \alpha \sum_{k=0}^{n_{\alpha}-1}  {\rm diag} \{\Psi(k) \otimes I_{p}, I_{Np} \}  F(X(k),k).
\end{align}
 For $t \geq 0$, $t \in \mathcal{R}$, define $X^{\alpha}(\cdot)$ as  $X^{\alpha}(t)=X(n)$ for $t \in [(n-n_{\alpha})\alpha, (n-n_{\alpha}+1)\alpha)$ (for details, see \cite{ky1}).

\begin{theorem} \label{th:1}  % THEOREM 1
	Let (A1)--(A7) hold. Let $X^{\alpha}(n)$ be generated  by (\ref{a}), (\ref{w}) and (\ref{partial}), with $\alpha_{i}(n)=\beta_{i}(n)=\alpha > 0$. Let $w^{\alpha}_{i}(0)=w^{\alpha}_{i,0}$, $e_{i}(0)=e_{i,0}$ $\in {\rm span} \{\phi(S) \}$. Define $X^{\alpha}(0)$  by $\lim_{\alpha \to 0} X_{0}^{\alpha}=[\theta_{0}^{T} \cdots \theta_{0}^{T} w_{1,0}^{T} \cdots w_{N,0}^{T} ]^{T}$. Then $X^{\alpha}(\cdot)$ is tight and  converges weakly to a process $X^{\alpha}(\cdot)=[\theta(\cdot)^{T} \cdots \theta(\cdot)^{T} w_{1}(\cdot)^{T} \cdots w_{N}(\cdot)^{T}]^{T}$, where $\theta( \cdot), w_{1} (\cdot), \ldots, w_{N}(\cdot)$ satisfy the following ODEs
	\begin{equation} \label{ode1}
	\dot{\theta}=\sum_{j=1}^{N} \bar{\psi}_{j} q_{j} \bar{g}_{j}(\theta,w_{j}), \;\;\;
	\dot{w_{i}}= \bar{k}_{i}(\theta,w_{i}),
	\end{equation}
	%where $\bar{F}(x)=[\bar{F}_{1}(x)^{T} \cdots  \bar{F}_{N}(x)^{T}]^{T}$, $\bar{F}_{i}(x)=[\bar{g}_{i}(\theta,w)^{T} \vdots \bar{k}_{i}(\theta,w)^{T}]^{T}$,
	$i=1, \ldots, N$, with initial conditions $\theta_{0}, w_{1,0}, \ldots, w_{N,0}$.
	%($\bar{g}_{i}(\theta,w)$ and $\bar{k}_{i}(\theta,w)$ are defined in  (\ref{barg}), (\ref{bark})).
	\par
	Moreover, for any integers $n'_{\alpha}$ such that $\alpha n'_{\alpha} \to \infty$ as $\alpha \to 0$, there exist positive numbers $\{T_{\alpha}\}$ with $T_{\alpha} \to \infty$ as $\alpha \to 0$, such that for any $\epsilon > 0$
	\begin{equation} \label{pnalpha1}
	%\lim \sup_{\alpha \to 0} P((\theta_{i}(n'_{\alpha}+k) \notin  N_{\epsilon}(\bar{\Sigma}_{\theta}) \; {\rm for} \;\ {\rm some} \;\ k \in [0, T_{\alpha}/\alpha])=0,
	\lim \sup_{\alpha \to 0} P\{X^{\alpha}(n'_{\alpha}+k) \notin  N_{\epsilon}(\bar{\Sigma}) \}=0
	\end{equation}
 for some  $ k \in [0, T_{\alpha}/\alpha]$,
 where $N_{\epsilon}( \cdot)$ denotes the $\epsilon$-neighborhood, while $\bar{\Sigma}= \bar{\Sigma}_{\bar{\theta}} \times \cdots \bar{\Sigma}_{\bar{\theta}} \times \bar{\Sigma}_{\bar{w}_{1}} \times \cdots \bar{\Sigma}_{\bar{w}_{N}}$  is the set of points $\bar{\theta}, \ldots, \bar{\theta},  \bar{w}_{1},  \ldots , \bar{w}_{N} $ satisfying
	\begin{equation} \label{limitp1}
	\sum_{j=1}^{N} \bar{\psi}_{j} q_{j} G_{j}^{T}  \bar{w}_{j}=0,  \;\;\;
	G_{i} \bar{\theta} +b_{i} -H_{i} \bar{w}_{i}=0, 	
	%&\sum_{i=1}^{N} \bar{\psi}_{i} G_{i}^{T}  \bar{w}_{i}=0, &
	%& \Theta_{opt}={\rm arg} \min_{\theta} \{ J(\theta) \}= {\rm arg} \min_{\theta} \{ \sup_{w} \tilde{J}(\theta,w) \} & \label{tetaopt}, \\
	%& w_{opt}= {\rm arg} \max_{w \in {\rm span}\phi(S)} \{\inf_{\theta} \tilde{J}(\theta,w)\} & \label{wopt}.
	\end{equation}
	$i=1, \ldots, N$, where $G_{i}=  \Phi^{T} \Xi_{i} (P^{(\lambda_{i})} - I) \Phi$, $b_{i}=\Phi^{T}  \Xi_{i} r_{\pi}^{(\lambda_{i})}$, $r_{\pi}^{(\lambda_{i})}$ is a constant $M$-vector in the affine function $T^{(\lambda_{i})}( \cdot)$, while $H_{i}=  \Phi^{T} \Xi_{i} \Phi$.
	%\begin{equation}+
	%J(\theta)=\sum_{i=1}^{N} \bar{\psi}_{i} J_{i}(\theta),
	%\end{equation}
	%\begin{equation}
	%\tilde{J}(\theta,w)= \sum_{i=1}^{N} \bar{\psi}_{i} \tilde{J}_{i}(\theta,w)
	%\end{equation}
	% with $\tilde{J}_{i}(\theta,w)=[\langle \Phi w, T^{(\lambda_{i})} \Phi \theta - \Phi \theta \rangle_{\xi_{i}}-\frac{1}{2} \|\Phi w \|^{2}_{\xi_{i}}]$
	%(for  $x=[x_{1} \cdots x_{N}]^{T} $,$y=[y_{1} \cdots y_{N}]^{T}$,  $\xi_{i}=[\xi_{i,1} \cdots \xi_{i,N}]^{T}$, $\langle x,y \rangle_{\xi_{i}}$ denotes the weighted scalar product $\langle x,y \rangle_{\xi_{i}}= \sum_{j=1}^{N} %x_{j} y_{j} \xi_{i,j}$).
\end{theorem}

{\bf Proof:}
%The proof is based on the application of the general results from \cite{ky1,ky,weak}, which correspond to a consensus based stochastic approximation scheme slightly different from (\ref{alg}). It includes verification of the the underlying basic assumptions from \cite{ky1} in light of the results presented in the preliminary part of this Section \cite{yu_off_policy_2017}.
\emph{Part 1.} Iterating (\ref{alg}) back, one obtains
%\begin{eqnarray} \label{iterate}
%& X_{n+1}=\tilde{\Phi}(n|0)X_{0}+ \varepsilon \sum_{k=0}^{n_{\varepsilon}-1} \tilde{\Psi}(n|k) F(Y_{k}, %\xi_{k})+ & \nonumber \\
%& +\varepsilon \sum_{k=n_{\varepsilon}}^{n} \tilde{\Psi}(n|k) F(Y_{k}, \xi_{k}),
%\end{eqnarray}
\begin{align} \label{iterate}
X(n+1)=& X_{0}^{\alpha} + \alpha \sum_{k=n_{\alpha}}^{n}{\rm diag} \{\Psi(k) \otimes I_{p}, I_{Np} \} F(X(k),k)+ \alpha \varrho(n)  \nonumber \\ & +{\rm diag} \{[\Psi(n|0) - \Psi(n_{\alpha}|0)] \otimes I_{p}, I_{Np} \}   X^{\alpha}_{0},
\end{align}
where
$\varrho(n)= \sum_{k=0}^{n} {\rm diag} \{[\Psi(n|k)-\Psi(k)]\otimes I_{p}, I_{Np} \}   F(X(k),k).$
At this point it is essential to verify the basic assumptions from \cite[Theorem~3.1]{ky1}. Using the preliminary part of this Section, we conclude that Lemma~\ref{lemma:1}, together with the results from \cite{yu_off_policy_2017}, imply that the assumptions C(3.2) and C(3.3') from Section~3 in \cite{ky1} are satisfied. Therefore,
 $\sup_{\alpha, n \geq n_{\alpha}} \frac{1}{\alpha^{2}} E \{ |X(n+1)-X(n)|^{2} \} < \infty$ and $\{\frac{1}{\alpha} |X(n+1)-X(n)|, n \geq n_{\alpha} \}$ is uniformly integrable,  $\{X^{\alpha}(\cdot) \}$ is tight and the limit paths Lipschitz continuous \cite[Theorem~3.1, Part~1]{ky1}.
 \par
The  asymptotic mean ODE (\ref{ode1}) follows, according to \cite{ky1}, from
\begin{align}
M_{f}(t)=f(X(t))-f(X(0)) +\int_{0}^{t}f'_{X}(X(s)) {\rm diag} \{\bar{\Psi} \otimes I_{p}, I_{Np} \}  \bar{F}(X(s))ds,
\end{align}
where $t$ is continuous time and $f(\cdot)$ a real valued function  with compact support and continuous second derivatives. Applying the Skorokhod embedding to the limit process $X^{\alpha}(\cdot) \rightarrow X(\cdot)$, one can show that $M_{f}(t)$ is a continuous martingale \cite{ky1}. Consequently, $M_{f}(t)=0$, having in mind that $X(\cdot)$ is Lipschitz continuous and that $M_{f}(0)=0$. This implies that $\dot{X}={\rm diag} \{\bar{\Psi} \otimes I_{p}, I_{Np} \} \bar{F}(X)$. By  Lemma~\ref{lemma:2} and (A2)--(A5), all the rows of $\bar{\Psi}$ are equal. It follows that the $p$-dimensional vector components of $\Theta$ must be equal, \emph{i.e.}, we obtain that $\Theta(\cdot)$ is in the form $\Theta(\cdot)=[\theta(\cdot)^{T} \cdots \theta(\cdot)^{T}]$, and that $\theta(\cdot)$ satisfies the first ODE from  (\ref{ode1}). The remaining ODEs related to $w_{i}$ immediately follow \cite[Theorem~8.2.2]{ky}.

\emph{Part 2.} In order to study the limit set of the ODE (\ref{ode1}),
%\begin{align}
%E=\cap_{\tau \geq 0}{\rm cl} \{ \theta(t), w_{1}(t), \ldots, w_{N}(t)|&\theta(0), w_{1}(0), \ldots, w_{N}(0) \nonumber \\ &\in \mathcal{R}^{(N+1)p}, t \geq \tau\},
%\end{align}
%where ${\rm cl} \{ \cdot \}$ denotes the closure of a given set.
%Let $f^{t}(D)= \{x(t)|x(0) \in D\}$ for any $D \subset \mathcal{R}^{2p}$. As we are dealing with unconstrained estimates, in our case $E=\cap_{t \geq 0}f^{t}(\mathcal{R}^{2p})$, representing the largest invariant subset of %$\mathcal{R}^{2p}$.
%\par
we shall follow  \cite[Proposition~4.1]{yu_off_policy_2017}, and introduce the Lyapunov function
\begin{equation}
V(\theta,w_{1}, \ldots, w_{N})=\frac{1}{2} \|\theta-\bar{\theta}\|^{2}+\frac{1}{2} \sum_{i=1}^{N} q_{i} \bar{\psi}_{i} \|w_{i}-\bar{w}_{i}\|^{2},
\end{equation}
 where $\bar{\theta}$ and $\bar{w}_{i}$ are given by (\ref{limitp1}). We have directly
\begin{equation}
\dot{V}(\theta,w_{1}, \ldots, w_{N})= - \sum_{i=1}^{N} q_{i} \bar{\psi}_{i} \langle w_{i}-\bar{w}_{i}, H_{i} (w_{i}-\bar{w}_{i}) \rangle,
\end{equation}
where $\langle \cdot , \cdot \rangle$ denotes the scalar product. Reasoning as in \cite{yu_off_policy_2017}, we infer that for  $w_{i}(0) \in {\rm span} \{\phi(\mathcal{S}) \}$ the set $\bar{\Sigma} $ satisfies (\ref{limitp1}).
\par
The remaining steps of the proof are standard for the applied methodology (see \cite{yu_off_policy_2017} and \cite[Theorem~8.2.2]{ky}).
\hspace*{\fill}\QED
\par
To deal with D2 algorithm types, we define $X_{0}^{\alpha}$ and $X^{\alpha}(\cdot)$ in the same way as above, but with replacing ${\rm diag} \{(A(n) \otimes I_{p}), I_{Np} \}$ by
${\rm diag} \{(A(n) \otimes I_{p}), (A(n) \otimes I_{p}) \}$ in the corresponding equations.
\begin{theorem} \label{th:2} %THEOREM 2
	Let (A1)--(A7) hold. Let $X^{\alpha}(n)$ be generated by (\ref{a}), (\ref{w}) and (\ref{full}), with $\alpha_{i}(n)=\beta_{i}(n)=\alpha >0$, and let both $w^{\alpha}_{i}(0)=w^{\alpha}_{i,0}$ and $e_{i}(0)=e_{i,0}$ $\in {\rm span} \{\phi(\mathcal{S}) \}$. Define $X^{\alpha}(0)$ by $\lim_{\alpha \to 0} X_{0}^{\alpha}=[\theta_{0}^{T} \cdots \theta_{0}^{T} w_{0}^{T} \cdots w_{0}^{T}]^{T}$. Then $X^{\alpha}(\cdot)$ is tight and  converges weakly to a process $X^{\alpha}(\cdot)=[\theta(\cdot)^{T} \cdots \theta(\cdot)^{T} w(\cdot)^{T} \cdots w(\cdot)^{T}]^{T}$, where $\theta(\cdot)$ and $w(\cdot)$ satisfy the following ODE:
	\begin{equation} \label{ode3}
	\left[ \begin{BMAT}{c}{cc} \dot{\theta} \\ \dot{w} \end{BMAT} \right]= \sum_{i=1}^{N} \bar{\psi}_{i} q_{i} \left[ \begin{BMAT}{c}{cc} \bar{g}_{i}(\theta,w) \\ \bar{k}_{i}(\theta,w) \end{BMAT} \right]
\end{equation}
 with initial conditions $\theta_{0}$ and $w_{0}$.
	%($\bar{g}_{i}(\theta,w)$ and $\bar{k}_{i}(\theta,w)$ are defined in  (\ref{barg}), (\ref{bark})).
	\par
	Moreover, for any integers $n'_{\alpha}$ such that $\alpha n'_{\alpha} \to \infty$ as $\alpha \to 0$, there exist positive numbers $\{T_{\alpha}\}$ with $T_{\alpha} \to \infty$ as $\alpha \to 0$ such that for any $\epsilon > 0$
	\begin{align} \label{pnalpha3}
	%\lim \sup_{\alpha \to 0} P((\theta_{i}(n'_{\alpha}+k), w_{i}(n'_{\alpha}+k)) \notin  N_{\epsilon}(\Theta^{*} \times W^{*}) \; {\rm for} \;\ {\rm some} \;\ k \in [0, T_{\alpha}/\alpha])=0,
	\lim \sup_{\alpha \to 0} P \{ &\left[ \begin{BMAT}{c}{cc} \theta_{i}^{\alpha} (n'_{\alpha}+k) \\  w_{i}^{\alpha}(n'_{\alpha}+k)  \end{BMAT} \right] \notin  N_{\epsilon}(\bar{\Sigma}) \}
	\end{align}
for some $ k \in [0, T_{\alpha}/\alpha]\}=0$,	$i=1, \ldots, N$, where $\bar{\Sigma}=\bar{\Sigma}_{\theta} \times \bar{\Sigma}_{w}$  is the set of points $\bar{x}=[\bar{\theta}^{T}  \bar{w}^{T}]^{T} \in \mathcal{R}^{2p}$ satisfying
	\begin{equation} \label{limitp3}
	\bar{G} \bar{\theta} +\bar{b} -\bar{H} \bar{w}=0,  \;\;\;\;\;
	\bar{G}^{T}  \bar{w}=0,
	%& \Theta_{opt}={\rm arg} \min_{\theta} \{ J(\theta) \}= {\rm arg} \min_{\theta} \{ \sup_{w} \tilde{J}(\theta,w) \} & \label{tetaopt}, \\
	%& w_{opt}= {\rm arg} \max_{w \in {\rm span}\phi(S)} \{\inf_{\theta} \tilde{J}(\theta,w)\} & \label{wopt}.
	\end{equation}
	where $\bar{G}= \sum_{i=1}^{N} \bar{\psi}_{i}q_{i} \Phi^{T} \Xi_{i} (P^{(\lambda_{i})} - I) \Phi$, $\bar{b}=\Phi^{T} \sum_{i=1}^{N} \bar{\psi}_{i} q_{i} \Xi_{i} r_{\pi}^{(\lambda_{i})}$, $r_{\pi}^{(\lambda_{i})}$ is a constant $M$-vector in the affine function $T^{(\lambda_{i})}( \cdot)$, while $\bar{H}=  \sum_{i=1}^{N} \bar{\psi}_{i} q_{i} \Phi^{T} \Xi_{i} \Phi$.
	%\begin{equation}
	%J(\theta)=\sum_{i=1}^{N} \bar{\psi}_{i} J_{i}(\theta),
	%\end{equation}
	%\begin{equation}
	%\tilde{J}(\theta,w)= \sum_{i=1}^{N} \bar{\psi}_{i} \tilde{J}_{i}(\theta,w)
	%\end{equation}
	% with $\tilde{J}_{i}(\theta,w)=[\langle \Phi w, T^{(\lambda_{i})} \Phi \theta - \Phi \theta \rangle_{\xi_{i}}-\frac{1}{2} \|\Phi w \|^{2}_{\xi_{i}}]$
	%(for  $x=[x_{1} \cdots x_{N}]^{T} $,$y=[y_{1} \cdots y_{N}]^{T}$,  $\xi_{i}=[\xi_{i,1} \cdots \xi_{i,N}]^{T}$, $\langle x,y \rangle_{\xi_{i}}$ denotes the weighted scalar product $\langle x,y \rangle_{\xi_{i}}= \sum_{j=1}^{N} %x_{j} y_{j} \xi_{i,j}$).
\end{theorem}

{\bf Proof:}
 The proof follows closely the proof of Theorem~\ref{th:1}.
In order to analyze the limit set of (\ref{ode3}), we introduce the Lyapunov function
%\begin{equation}
$V(\theta,w)=\frac{1}{2} \|\theta-\bar{\theta}\|^{2}+\frac{1}{2}  \|w-\bar{w}\|^{2}$,
%\end{equation}
 where $\bar{\theta}$ and $\bar{w}$ are given by (\ref{limitp3}). We have directly
%\begin{equation}
$\dot{V}(\theta,w)=  - \langle w-\bar{w}, \bar{H} (w-\bar{w}) \rangle$,
%\end{equation}
wherefrom the result follows.
\hspace*{\fill}\QED
%%\emph{D1-GTD2($\lambda$), $\varepsilon=0$}
%when
%$\alpha_{i}(n)=\alpha$ and  $\beta_{i}(n)=\beta$, but $\beta >> \alpha$.
%\par

The next theorem deals with two-time-scale versions of the algorithms.
\begin{theorem} \label{th:3}   %THEOREM 3
Let (A1)--(A7) hold. Let $X^{\alpha,\beta}(n)$ be generated by (\ref{a}), (\ref{w}) and (\ref{partial}), with $\alpha_{i}(n)=\alpha > 0$, $\beta_{i}(n)=\beta > 0$, $\beta >> \alpha$, and let both $w^{\alpha,\beta}_{i}(0)=w^{\alpha,\beta}_{i,0}$ and $e_{i}(0)=e_{i,0}$ $\in {\rm span} \{\phi(S) \}$. Define $X^{\alpha,\beta}(0)$ by $\lim_{\beta \to 0, \alpha/\beta \to 0} X_{0}^{\alpha,\beta}=[\theta_{0}^{T} \cdots \theta_{0}^{T} w_{1,0}^{T} \cdots w_{N,0}^{T}]^{T}$. Then $X^{\alpha,\beta}(\cdot)$ is tight and converges weakly at the fast time scale to a process $W(\cdot)= [w_{1}(\cdot)^{T} \cdots w_{N}(\cdot)^{T} ]^{T}$ generated by
\begin{equation} \label{ode2}
 \dot{w}_{i}=\bar{k}_{i}(\theta_{i},w_{i}),
 \end{equation}
for any given $\theta_{1}, \ldots, \theta_{N}$, with $w_{i,0} \in {\rm span} \{\phi(S)\}$, $i=1, \ldots, N$, and at the slow time scale to $\Theta(\cdot)=[\theta(\cdot)^{T} \cdots \theta(\cdot)^{T}]^{T}$, where
\begin{equation} \label{odeg}
\dot{\theta}=\sum_{i=1}^{N} \bar{\psi}_{i} q_{i} \bar{g}_{i}(\theta, \bar{w}_{i}(\theta)),
\end{equation}
with the initial condition $\theta_{0}$, where $\bar{w}_{i} (\theta)$ is the unique solution (w.r.t. $w_{i}$) of the equation
\begin{equation}
\bar{k}_{i}(\theta, w_{i})=G_{i} \theta + b_{i} -H_{i} w_{i}=0.
\end{equation}

%($\bar{g}_{i}(\theta,w)$ and $\bar{k}_{i}(\theta,w)$ are defined in  (\ref{barg}), (\ref{bark})).
\par
Moreover, for any integers $n'_{\alpha,\beta}$ such that $\alpha n'_{\alpha,\beta} \to \infty$ as $\alpha \to 0$, there exist positive numbers $\{T_{\alpha,\beta}\}$ with $T_{\alpha,\beta} \to \infty$ as $(\beta, \alpha/\beta) \to 0$ such that for any $\epsilon > 0$
\begin{align} \label{pnalpha2}
%\lim \sup_{\alpha \to 0} P((\theta_{i}(n'_{\alpha}+k), w_{i}(n'_{\alpha}+k)) \notin  N_{\epsilon}(\Theta^{*} \times W^{*}) \; {\rm for} \;\ {\rm some} \;\ k \in [0, T_{\alpha}/\alpha])=0,
\lim \sup_{\beta \to 0, \alpha/\beta \to 0} P\{&(\theta^{\alpha,\beta}_{i}(n'_{\alpha,\beta}+k)) \notin  N_{\epsilon}(\bar{\Sigma}_{\bar{\theta}})\} =0
\end{align}
for some $k \in [0, T_{\alpha,\beta}/\alpha]$, $i=1, \ldots, N$, where  $\bar{\Sigma}_{\bar{\theta}}$  is the set of points $\bar{\theta} \in \mathcal{R}^{p}$ defined by $\sum_{i=1}^{N} \bar{\psi}_{i} q_{i} G_{i}^{T} \bar{w}_{i}(\bar{\theta})=0$.

% satisfying
%\begin{equation} \label{limitp}
%\bar{G} \bar{\theta} +\bar{g} -\bar{H} \bar{w}=0,  \;\;\;\;\;
% \bar{G}^{T}  \bar{w}=0,
%& \Theta_{opt}={\rm arg} \min_{\theta} \{ J(\theta) \}= {\rm arg} \min_{\theta} \{ \sup_{w} \tilde{J}(\theta,w) \} & \label{tetaopt}, \\
%& w_{opt}= {\rm arg} \max_{w \in {\rm span}\phi(S)} \{\inf_{\theta} \tilde{J}(\theta,w)\} & \label{wopt}.
%\end{equation}
%where $\bar{G}= \sum \bar{\psi}_{i} \Phi^{T} \Xi_{i} (P^{(\lambda_{i})} - I) \Phi$, $\bar{g}=\Phi^{T} \sum_{i} \Xi_{i} r_{\pi}^{(\lambda_{i})}$, $r_{\pi}^{(\lambda_{i})}$ is a constant $p$-vector in the affine function $T^{(\lambda_{i})}( \cdot)$, while $\bar{H}=  \sum_{i=1}^{N} \bar{\psi}_{i} \Phi^{T} \Xi_{i} \Phi$.
%\begin{equation}
%J(\theta)=\sum_{i=1}^{N} \bar{\psi}_{i} J_{i}(\theta),
%\end{equation}
%\begin{equation}
%\tilde{J}(\theta,w)= \sum_{i=1}^{N} \bar{\psi}_{i} \tilde{J}_{i}(\theta,w)
%\end{equation}
% with $\tilde{J}_{i}(\theta,w)=[\langle \Phi w, T^{(\lambda_{i})} \Phi \theta - \Phi \theta \rangle_{\xi_{i}}-\frac{1}{2} \|\Phi w \|^{2}_{\xi_{i}}]$
%(for  $x=[x_{1} \cdots x_{N}]^{T} $,$y=[y_{1} \cdots y_{N}]^{T}$,  $\xi_{i}=[\xi_{i,1} \cdots \xi_{i,N}]^{T}$, $\langle x,y \rangle_{\xi_{i}}$ denotes the weighted scalar product $\langle x,y \rangle_{\xi_{i}}= \sum_{j=1}^{N} %x_{j} y_{j} \xi_{i,j}$).
\end{theorem}

{\bf Proof:}
The proof can be derived using \cite[Section~3]{ky1}, proof of Theorem~\ref{th:1} and the general results on weak convergence of two-time-scale stochastic approximation algorithms \cite[paragraph~8.6]{ky}, \cite{borkar1997stochastic,borkar}.
The first part of the proof is analogous to the first part of the proof of Theorem~\ref{th:1}.  
As far as the invariant set of the mean ODEs is concerned, for the fast time scale we have (\ref{ode2}), since  $(\alpha/\beta) \bar{g}_{i}(\theta,w)$ is negligible when $\beta, \alpha/\beta \to 0$. As for any given $\theta$ there is a unique solution $\bar{w}_{i}(\theta)$ to the linear equation  $\bar{k}_{i}(\theta,w_{i})=0$,  $w_{i} \in {\rm span} \{\phi(S) \}$,  we have (\ref{odeg}) for the slow time scale.
\par
In order to prove (\ref{pnalpha2}), we introduce the Lyapunov function
%\begin{equation}
$V(\theta)=\sum_{i=1}^{N} \bar{\psi}_{i} q_{i} J_{i}(\theta)$,
%\end{equation}
using (\ref{j}), so that
%\begin{equation}
$\dot{V}(\theta)=
-\left\|\sum_{i=1}^{N} \bar{\psi}_{i} q_{i} \bar{g}_{i}(\theta, \bar{w}_{i}(\theta)) \right\|^{2}$.
%\end{equation}
It follows that $\dot{V}(\theta)=0$ if $\theta \in \bar{\Sigma}_{\theta}$; if $\theta \notin \bar{\Sigma}_{\theta}$, then, $\sum_{i=1}^{N} \bar{\psi}_{i} q_{i} \bar{g}_{i}(\theta, \bar{w}_{i}(\theta))\neq 0$ and hence $\dot{V}(\theta) < 0$.
\hspace*{\fill}\QED

\begin{remark}
Algorithm GTD2($\lambda$) has been originally proposed in the form of an one-time-scale algorithm \cite{sutton2009fast}; in \cite{yu_off_policy_2017}, it has been defined and analyzed as a two-time-scale algorithm. Algorithm TDC($\lambda$) has been proposed and analyzed only as a two-time-scale algorithm \cite{sutton2009fast,yu_off_policy_2017}. In general, the two-time-scale setting is natural, having in mind properties of $w$ as a faster auxiliary variable. By our experience, the algorithms of GTD2-type can be efficient in both cases, while those of TDC-type perform well only in the two-time-scale case. See the simulation section for a performance comparison.
\end{remark}
\begin{remark}
Algorithms with or without consensus w.r.t. $w_{i}$ have, in general, different convergence points for $\theta$. Consider, for example,
algorithms D1-GTD2$(\lambda)$ and D2-GTD2($\lambda$). If $\bar{\theta}$ denotes a convergence point, it can be easily seen from the Theorems~\ref{th:1} and \ref{th:2} that, in the first case, $\bar{\theta}$ follows from
$\sum_{i} \bar{\psi}_{i} G_{i}^{T} H_{i}^{-1} (G_{i} \bar{\theta}+b_{i})=0$, and, in the second, from $\bar{G^{T}} \bar{H}^{-1}(\bar{G} \bar{\theta} + \bar{b})=0$ (assuming that $H_{i}$ and $\bar{H}$ are nonsingular). The solutions are equal in the case of equal $\lambda$-parameters and equal behavior policies for all the agents. Notice that in the case of D1-GTD2$(\lambda)$ $\theta$ corresponds to the strictly optimal solution w.r.t. (\ref{j}). However, D2-GTD2$(\lambda)$ is practically more favorable in the cases of significantly different behavior policies, reducing the estimation variance (see Section~\ref{sec:sim} for an example). In general, consensus on $w$ may cause somewhat slower response, more visible in the one-time-scale setting. 
The two-time-scale setting allows getting faster response and lower variance for $\theta$.
\end{remark}
\begin{remark}
Following \cite{yu_off_policy_2017}, it is possible to obtain convergence results for diminishing step-sizes converging to zero at a rate lower than $1/n$. We have selected constant step-sizes motivated by practical applications to slowly time-varying cases. It is also possible to extend the results and to prove convergence w.p.1 at the expense of additional constraints, see, \emph{e.g.},  \cite{ky1,yu_off_policy_2017}.
\end{remark}
\begin{remark}
It is possible to generalize the problem setting by assuming that the quadruplets $\mathcal{Q}_{i}$ have the same state and action spaces, but that the probabilities characterizing the environment and the reward distribution are agent dependent ($p_{i}(s'|a,s)$ and $q_{i}(\cdot|s',a,s)$). The optimization problem from (\ref{j}) becomes \emph{multicriterial}, providing the figure of merit of a given target policy applied in parallel in different environments. The above derivations basically hold; however, the interpretation of the results is not as straightforward as above.
\end{remark}

\section{Discussion} \label{sec_discussion}

\subsection{Constrained Algorithms} \label{subsec:constr}
It is possible to formulate \emph{constrained versions} of all the  proposed algorithms and to prove their weak convergence following the methodology developed for the single agent case \cite{yu_off_policy_2017}. Formally, the constrained form of the algorithms is obtained  by applying projections $\Pi_{B_{\theta}}\{ \}$ and $\Pi_{B_{w}} \{ \cdot \}$  of the right hand sides of (\ref{a}), (\ref{w}) and \eqref{TDC} on predefined constraint sets $B_{\theta}$ and $B_{w}$, respectively, w.r.t. $\| \cdot \|_{2}$ \cite{yu_off_policy_2017}.
%\begin{align}
%\theta'_{i}(n)=&  \Pi_{B_{\theta}}\{\theta_{i}(n)+ \alpha_{i}(n) q_{i} \rho_{i}(n)(\phi(S_{n}) \nonumber \\ &-\gamma(n+1) \phi(S_{n+1})) e_{i}(n)^{T}w_{i}(n)\} \label{ac}  \\
%w'_{i}(n)=&\Pi_{B_{w}}\{ w_{i}(n)+ \beta_{i}(n)(e_{i}(n) \delta_{i}(v_{\theta_{i}(n)};n) \nonumber \\ &-\phi(S_{n}) \phi(S_{n})^{T} w_{i}(n))\}, \label{wc}
%\end{align}
Notice also that a general analysis of constrained distributed stochastic approximation algorithms is presented in \cite{ky,ky1}. Assumption~(A7) should be removed in this case.  %Also, one has to take into account that the asymptotic ODE's contain now the boundary reflection terms, influencing, in general, definition of their limit sets \cite{yu_off_policy_2017} and imposing, in some cases, additional constraints w.r.t. $B_{\theta}$ and $B_{w}$ (see e.g. Lemmas~3.1 and 3.2 from \cite{yu_off_policy_2017}).
%\subsection{Asymptotic Optimality w.r.t. $J(\Theta)$}

%It is possible to consider the problem of bias, considered as distance of a convergence point from the reference estimate obtained in the case of on-policy learning. Having in mind numerous adaptable parameters characterizing the multi-agent network, it is possible to formulate a corresponding optimization task [da li da pomenemo ovde CPS rad].
\subsection{Asymptotic Convergence Rate: Covariance Reduction}
Consider D2-GTD2($\lambda$) in the light of \cite[Section~6]{ky1}. Define
\begin{equation}
U^{\alpha}(n)=\frac{Y^{\alpha}(n)-\bar{Y}}{\sqrt\alpha},
\end{equation}
where $Y^{\alpha}(n)$ follows from the global model   $Y^{\alpha}(n)=[x_{1}(n)^{T} \cdots x_{N}(n)^{T}]^{T}$, $x_{i}(n)=[\theta_{i}(n)^{T} w_{i}(n)^{T}]^{T}$ and $\bar{Y}=[\bar{x}^{T} \cdots \bar{x}^{T}]^{T}$, $\bar{x}=[\bar{\theta}^{T} \bar{w}^{T}]^{T}$, and assume it is tight for $n \geq N_{T}$. Define also
\begin{equation}
V^{\alpha}(n)= \sqrt \alpha \sum_{k=N'_{T}+n_{\alpha}+1}^{n}  (\Psi(k) \otimes I_{2p})  F^{Y}(\bar{Y},n),
\end{equation}
where $F^{Y}(\bar{Y},n)=[F^{Y}_{1}(\bar{Y},n)^{T} \cdots F^{Y}_{N}(\bar{Y},n)^{T}]^{T}$, $F_{i}^{Y}(\bar{Y},n)=$  $[q_{i} g_{i}(\bar{\theta}_{i},\bar{w}_{i},Z_{i}(n))^{T} \vdots $ $ q_{i}k_{i}(\bar{\theta}_{i},\bar{w}_{i}, Z_{i}(n))^{T}$ $+q_{i}e_{i}(n)^{T} \omega_{i}(n+1)]^{T}$ and $N'_{T} \geq N_{T}$.

Following \cite[Section~5.1]{ky1}, it is possible to show that when $X^{\alpha}(n)$ converges weakly to $X(\cdot)=[\theta(\cdot)^{T} \cdots \theta(\cdot)^{T} w(\cdot)^{T} \cdots w(\cdot)^{T}]^{T}$ (according to Theorem~\ref{th:2}), we have also weak convergence of $Y^{\alpha}(n)$ to $Y( \cdot)=[x(\cdot)^{T} \cdots x(\cdot)^{T}]^{T}$, as well as of $U^{\alpha}(n)$ and $V^{\alpha}(n)$ to $U(\cdot)= [u(\cdot)^{T} \cdots u(\cdot)^{T}]^{T}$ and  $V(\cdot)=[v(\cdot)^{T} \cdots v(\cdot)^{T}]^{T}$, respectively. It is also possible to show that vectors $u(\cdot)$ and $v(\cdot)$ asymptotically satisfy the following It${\rm \hat{o}}$ \emph{stochastic differential equation} (SDE)
\begin{equation} \label{sde}
du=Qu \, dt+d\varrho,
\end{equation}
where matrix $Q$ is the Jacobian matrix of $( \hat{\Psi} \otimes I_{2p}) \bar{F}^{Y}(\bar{Y})$ ($\bar{F}^{Y}(\bar{Y})$ follows from $F^{Y}(\bar{Y},n)$ in the same way as $\bar{F}(X)$ follows from $F(X,n)$ in the global model description) and  $\varrho(\cdot)$ a Wiener process satisfying
\begin{align}
\mathrm{cov} \{ \varrho(1) \}=\bar{R} =\sum_{k = -\infty}^{k = \infty} E \left\{ \left[\sum_{i=1}^{N} \psi_{i}(k) F^{Y}_{i}(\bar{x},k)\right] \left[\sum_{i=1}^{N} \psi_{i}(k) F^{Y}_{i}(\bar{x},k)\right] ^{T}\right\},  \nonumber
\end{align}
where $\psi_{1}(k), \ldots,  \psi_{N}(k)$ are the elements of each row of the row-stochastic time-varying random matrix $\Psi(k)$ ($E\{ \cdot\}$ is understood in the sense of the ergodic mean).
\par
The stationary covariance $R_{u} = \int_{0}^{\infty} e^{Qt} \bar{R} e^{Q^{T}t}dt$ can be taken as a measure of noise influence.
For the sake of clarity, we shall consider a very simple case assuming that $F^{Y}_{i}(\cdot)=F^{Y}(\cdot)$, ${\rm cov} \{ F^{Y}(\bar{Y},n)\}=R_{i}=R$ and  $p=1$. Then,
${\rm cov} \{ \varrho(1)\}=R \sum_{i=1}^{N} E\{ \psi_{i}(n)^{2} \}$.
In the case of no network, the SDE model has the same form (\ref{sde}), but with ${\rm cov} \{ \varrho(1) \} = R$. The advantage of the consensus based algorithm is obvious, having in mind that $\Sigma_{i=1}^{N} E\{ \psi_{i}(n)^{2} \} < 1$.
\begin{remark}
Variance reduction is one of the general problems in temporal difference algorithms \cite{sutton1998reinforcement,tvr,emph,emph_yu}. The above result shows that the consensus-based averaging may provide significant improvements of asymptotic covariance w.r.t. the single-agent case. In this sense, the ``denoising" phenomenon may represent one of the motivations for adopting a consensus-based approach to value function approximation  (see also the results from \cite{ky1,weak}). Of course, rigorous treatment of more general cases requires additional effort (see Section~\ref{sec:sim} for some examples).
\end{remark}
\subsection{Inter-Agent Communications and Network Design}
%The proposed distributed multi-agent algorithms can be considered as: 1) a tool for organizing coordinated actions of multiple agents contributing to the value function estimation and 2) a parallelization tool, allowing faster convergence, useful particularly in the problems with large dimensions. Notice that the in the first case the proposed algorithms can become a part of multi agent actor-critic schemes (see, \emph{e.g.}, \cite{zhang2018fully,basar2019}).
In general, the agents have specifically tailored behavior policies (including different ways of defining the local $\lambda$-parameters), having in mind that complementary exploration can contribute significantly to the overall rate of convergence. Factors in $\bar{\psi}_{i} q_{i}$, $i=1, \ldots, N$, allow placing more emphasis on selected agents.
%In this sense, it appears advisable to implement multi-step consensus within time intervals between successive observations \cite{aiss2}.
Generically, $q_{i}$ is chosen \emph{a priori}, while $\bar{\psi}_{i}$ depends solely on the network properties through the definition of matrix $A(n)$. There is a great flexibility from the point of view of \emph{network design}. For example, if one adopts that  $A(n)=A$, the problem reduces to the definition of a constant $N \times N$ matrix $A$ satisfying a given topology (defined by $A_{\mathcal{G}}$),
which provides  $\bar{\psi}_{i}=\frac{1}{N}$. Formally, one has to solve for $A$ the standard equation $\mathbf{1}^{T}
A=\mathbf{1}^{T}$, where $\mathbf{1}^{T}=[1 \cdots 1]^{T}$, which always has a
solution in our case
\cite{weak}.
Furthermore, the adopted algorithm formulation allows random matrices $A(n)$,
allowing treating \emph{communication dropouts} and
 \emph{asynchronous communications}. A detailed
analysis of this problem is given in \cite{weak} for \emph{broadcast }\emph{gossip}.

\subsection{Algorithms under Weak ISC}
Following a number of recent papers devoted to distributed value function estimation \cite{leehu2020,dwywj2020,cassano_sayed_linear_rate_ECC_2019,zhang_zavlanos_2019,basar2019,zhang2018fully}, it is possible to assume that the adopted ISC subsumes accessibility of all the states and actions by all the agents. This assumption may appear to be unrealistic for standard WSN's; however, the above results can be easily extended to this case.

Assume that a \emph{multi-agent system} is defined by the quadruplets $\bar{\mathcal{Q}}_{i}=\{\mathcal{S},\mathcal{A},P(s'|a,s),R_{i}(s',a,s) \}$, $i=1, \ldots, N$, where $\mathcal{S}=\prod_{i}\mathcal{S}_{i}$, $\mathcal{A}=\prod_{i}\mathcal{A}_{i}$ ($\mathcal{S}_{i}$ and $\mathcal{A}_{i}$ are finite local state and action spaces), tensor $P(s'|a,s)$ defines the global probabilities for all $s',s \in \mathcal{S}$ and $a \in \mathcal{A}$, while $ R_{i}(s',a,s)$ are the local random rewards with probability distributions $q_{i}(\cdot| s',a,s)$ depending, in general, on $i$. The global behavior policy is  $b(a|s)=\prod_{i} b_{i}(a|s)$  and the global target policy $\pi(a|s)=\prod_{i}\pi_{i}(a|s)$, $s',s \in \mathcal{S}$, $a \in \mathcal{A}$, where $b_{i}$ and $\pi_{i}$ are local behavior and target policies, respectively. The value function follows directly from (\ref{Bel0}), as well as its linear approximation. The steps leading to distributed algorithms are identical as above; formally one comes to (\ref{a}), (\ref{w}), \eqref{TDC}, (\ref{partial}) and (\ref{full}), where index $i$ remains only in the stochastic reward term $R_{i}(n)$. Weak convergence to consensus can be proved similarly as above. Notice only that the proposed algorithms provide an estimate of the global value function for the fictitious global random reward $\sum_{i} q_{i} \bar{\psi}_{i} R_{i}(s',a,s)$.

\section{Simulation Results} \label{sec:sim}

\begin{figure}
\begin{center}
\includegraphics[width=0.7\columnwidth,keepaspectratio,clip]{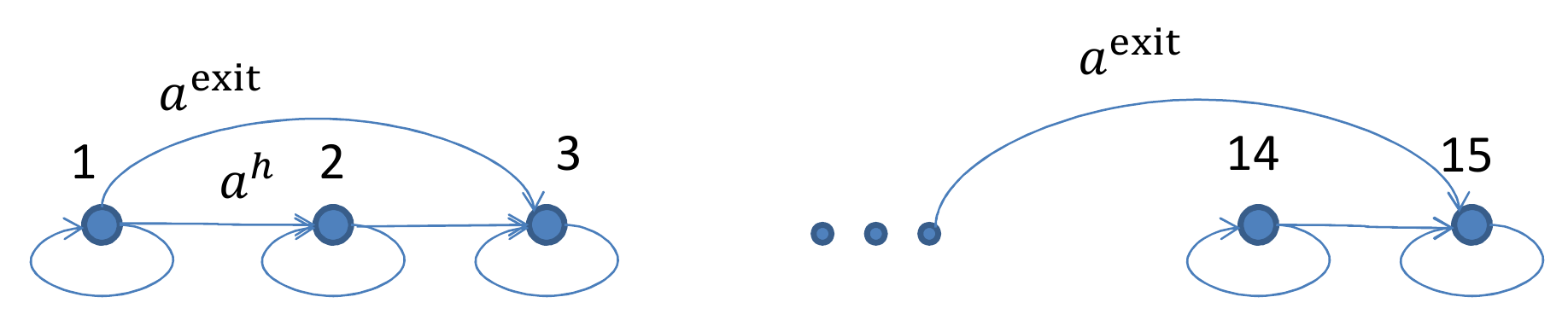}
\end{center}
\caption{Diagram of the simulated MDP.}
\label{chain}
\end{figure}

In this section we illustrate the main properties of the proposed algorithms by applying them to a version of the Boyan's chain, an environment frequently used in the literature, e.g. \cite{sutton2009fast,valcarcel2013cooperative,stankovic_ACC_2016}. The diagram of the underlying Markov chain is shown in Fig.~\ref{chain} \cite{stankovic_ACC_2016}.

The chain has 15 states with one absorbing state. We assume that $\gamma=0.85$. The chain can be interpreted as a decision making problem on a highway, with possibilities of exiting (using alternative roads). The policy which a driver can choose at each state is the probability of selecting the exit action $a^{\rm exit}$ at state $s$: $\pi(s,a^{\rm exit})$. The reward for exiting is $r(s,a^{\rm exit},s')=-4$ for all $s$ and $s'$ (can be interpreted as the consumed fuel), but the probability of staying in the same state (jammed) is fixed to 0.2. If we choose action $a^{h}$ (to stay on the highway) the reward is $r(s,a^h,s')=-1$ for all $s$ and $s'$, but the probability of staying in the same state grows with the state number as $1-\frac{1}{s}$, where $s$ is the state number. The \textit{target policy} is the stationary policy $\pi(s,a^{\rm exit})=0.8$. We assume that there are 10 agents with time-invariant communication graph, such that the agents communicate only with three randomly chosen neighbors, all taken with equal weights. The agents are only able to obtain 7-features Gaussian radial basis representations of the state vector as functions of distances to the states 1, 3, 5, 7, 9, 11 and 13 ($\phi_i(s)=e^{\frac{(s-z_i)^2}{2\sigma^2}}$, $i=1,...,7$, $z_i\in\{1,3,5,7,9,11,13\}$, with $\sigma^2=2$). Note that the chain has an absorbing state (it does not satisfy the conditions for convergence); hence we run the algorithms in multiple episodes by resetting the states back to 1 when the absorbing state is reached.

\begin{figure}
\begin{center}
\includegraphics[width=0.6\columnwidth]{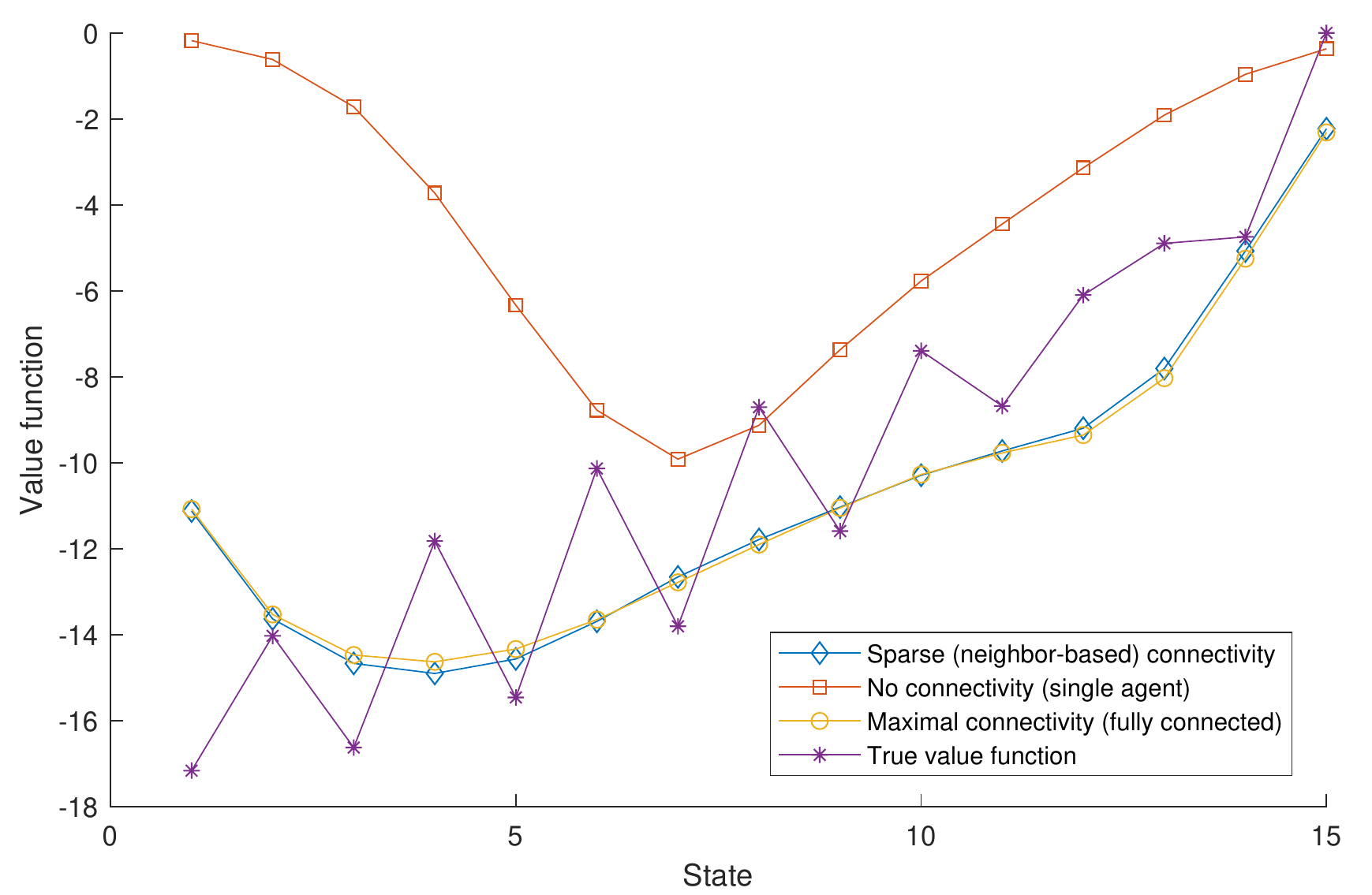}
\end{center}
\caption{Value function approximation obtained by one agent using D2-TDC($\lambda$) in which the agents have behavior policies such that they can individually visit only a subset of the states. True value function is shown using purple line. Different colors of the obtained approximations correspond to different network connectivity levels.}
\label{dtdc_value_exp2}
\end{figure}

In the first experiment, we demonstrate the case in which the agents, individually, are not able to estimate the value function due to their restrictive behavior policies; however, they are able to obtain convergent estimates of the value function using the proposed consensus algorithm. We assume that these policies are such that the agents can individually visit only a subset of the states, with the following agents' starting and stopping states $[(1,3),(2,4),(4,7),(5,15),(5,14),(3,14),(8,14),$ $(1,6),(5,10),(6,11)]$, i.e. the first agent always starts in state 1 and stops in state 3, and so on. Formally, we model this situation by assuming a possibility of choosing the third action (besides $a^h$ and $a^{\rm exit}$), which makes the current state absorbing. While visiting the allowed subsets of the states the agents have the following stationary behavior policies $[\pi_1(s,a^{\rm exit}), \pi_2(s,a^{\rm exit}),...,\pi_{10}(s,a^{\rm exit})]=[0.64, 0.75, 0.5, 0.81, 0.85, 0.8, 0.3, 0.55, 0.45, 0.6]$. In Fig.~\ref{dtdc_value_exp2} the value function approximations obtained by the agent 10 (which is only capable of visiting states from 6 to 11) using D2-TDC($\lambda$) algorithm, for $\lambda_i=0.5$, $i=1,...,10$, using constant step sizes $\alpha=0.3$ and $\beta=2$ (two time scales), are shown. The true value function is depicted using the purple line, while the other colors correspond to the obtained approximations assuming three different network connectivities: a) sparse, neighborhood based, connectivity introduced above, b) no connectivity (single agent case), and c) fully connected graph (all-to-all connectivity). Similar results for the final estimates for the described case are obtained for the rest of the proposed algorithms. It can be observed that, in the two connected cases, better approximation of the value function is obtained for the latter states, because the behavior policies of the agents are such that overall they visit these states more frequently (with higher probability), and, hence, they will have higher weights in the overall criterion \eqref{j}. Obviously, in the case in which there are no consensus-based collaborations, agent 10 is not capable to obtain good overall approximation.

\begin{figure}
\begin{center}
\includegraphics[width=0.6\columnwidth]{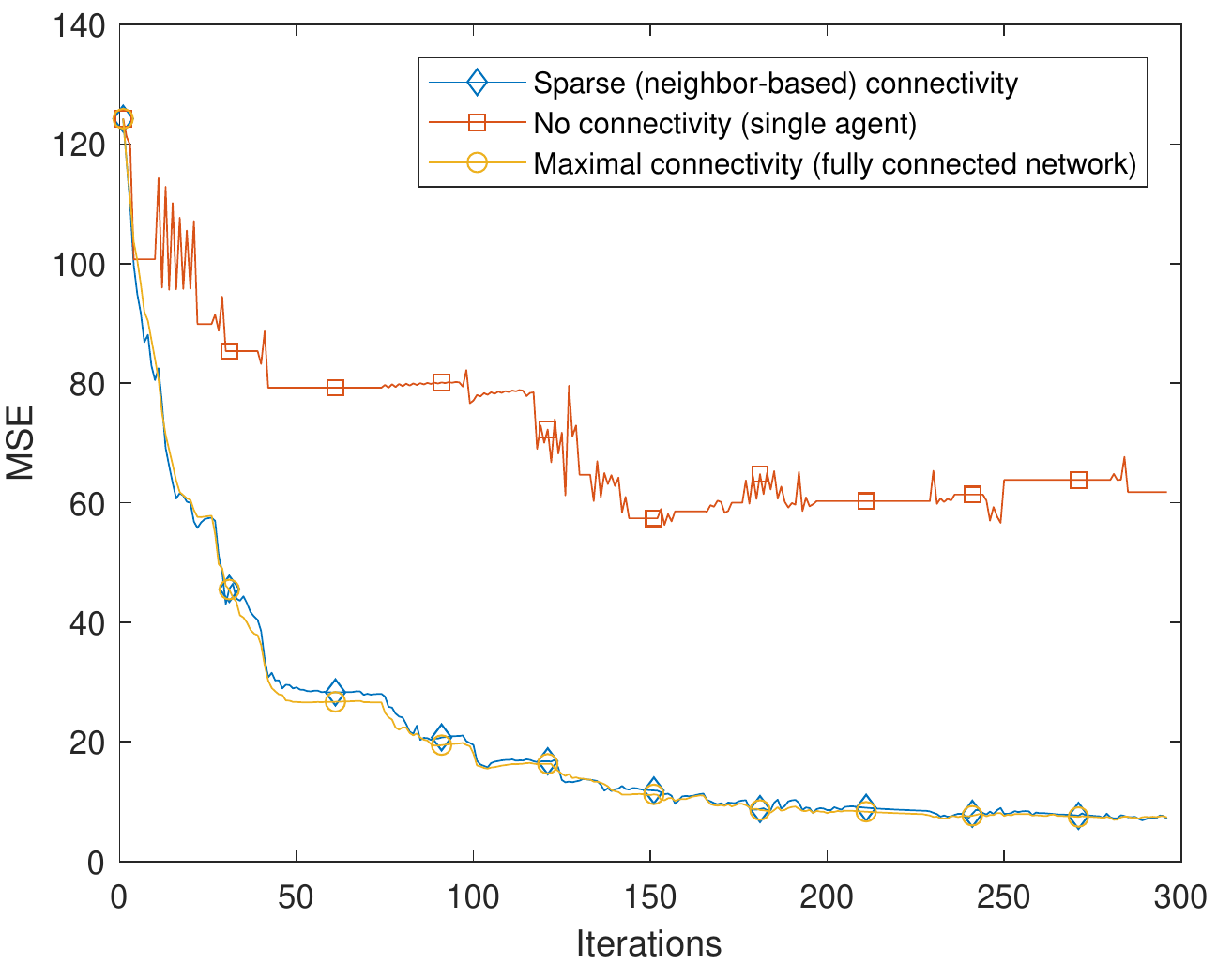}
\end{center}
\caption{Mean-square error of value function approximation obtained by one agent using D2-TDC($\lambda$), for different network connectivities, for the case in which the agents have behavior policies such that they can individually visit only a subset of states. Different colors of the curves correspond to different network connectivity levels.}
\label{fig_net_comp1}
\end{figure}

The benefit of the introduced consensus-based scheme can also be inferred from Fig.~\ref{fig_net_comp1}, where the mean-square error (MSE) of the value function approximation (averaged over all states), as a function of the number of iterations, is shown for the node 10, with the same algorithm as above. Different curves correspond to different network connectivity levels as described above.

\begin{figure}
\begin{center}
\includegraphics[width=0.6\columnwidth]{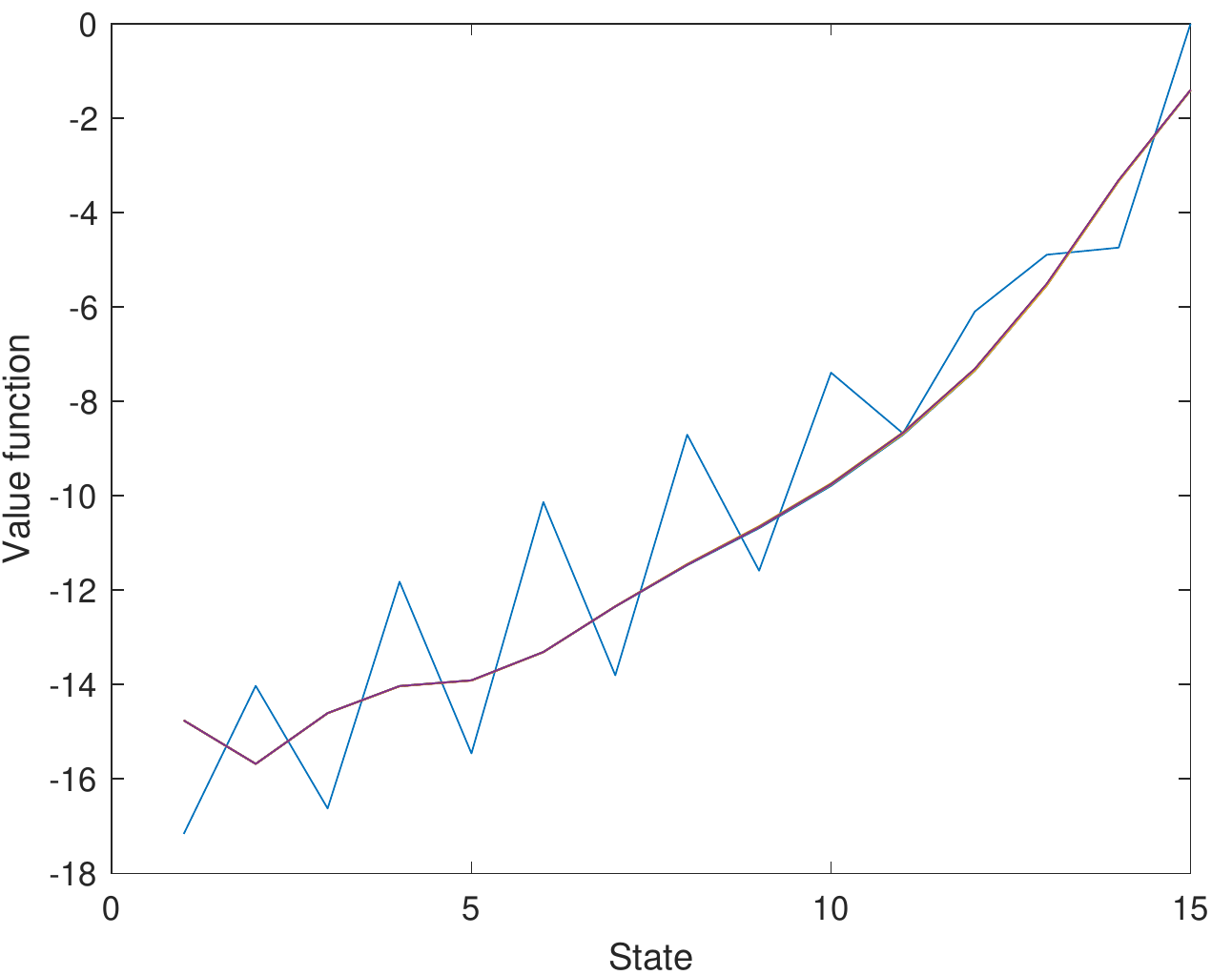}
\end{center}
\caption{Value function approximation obtained using D1-GTD2($\lambda$) in which all the agents have behavior policies such that they can visit all the states. True value function is shown using blue line. Different colors of the obtained approximations correspond to different agent's estimates so that it can be observed that the agents have practically achieved consensus.}
\label{dgtd2_value}
\end{figure}
\begin{figure}
\begin{center}
\includegraphics[width=0.6\columnwidth]{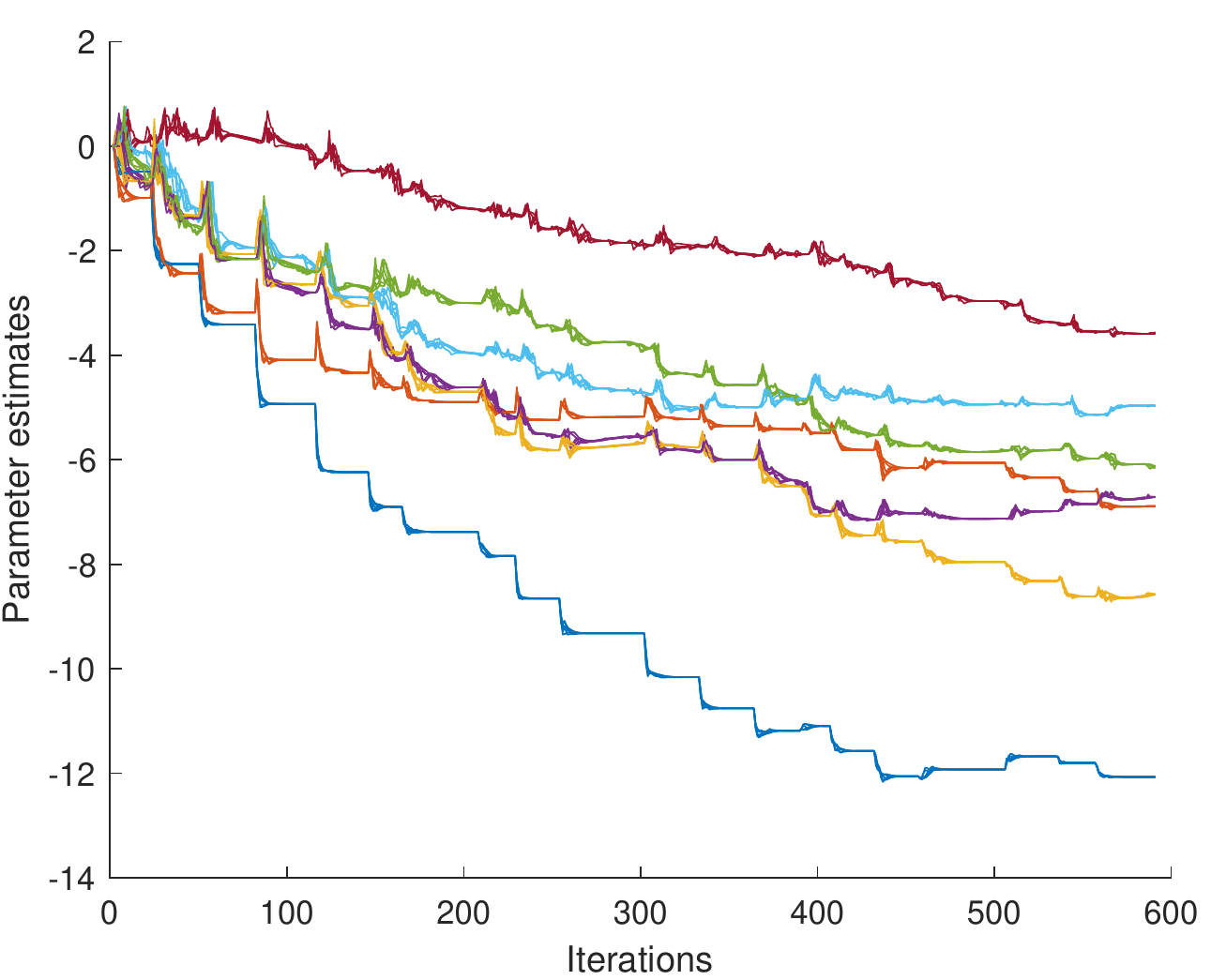}
\end{center}
\caption{Parameter estimates for all the agents using D1-GTD2($\lambda$) in the second experiment. Each color corresponds to a different parameter (with 7 parameters total). Curves with the same color (which are very close to each other due to the consensus) correspond to the estimates of the same parameter for all the agents.}
\label{dgtd2_parameters}
\end{figure}
\begin{figure}
\begin{center}
\includegraphics[width=0.6\columnwidth]{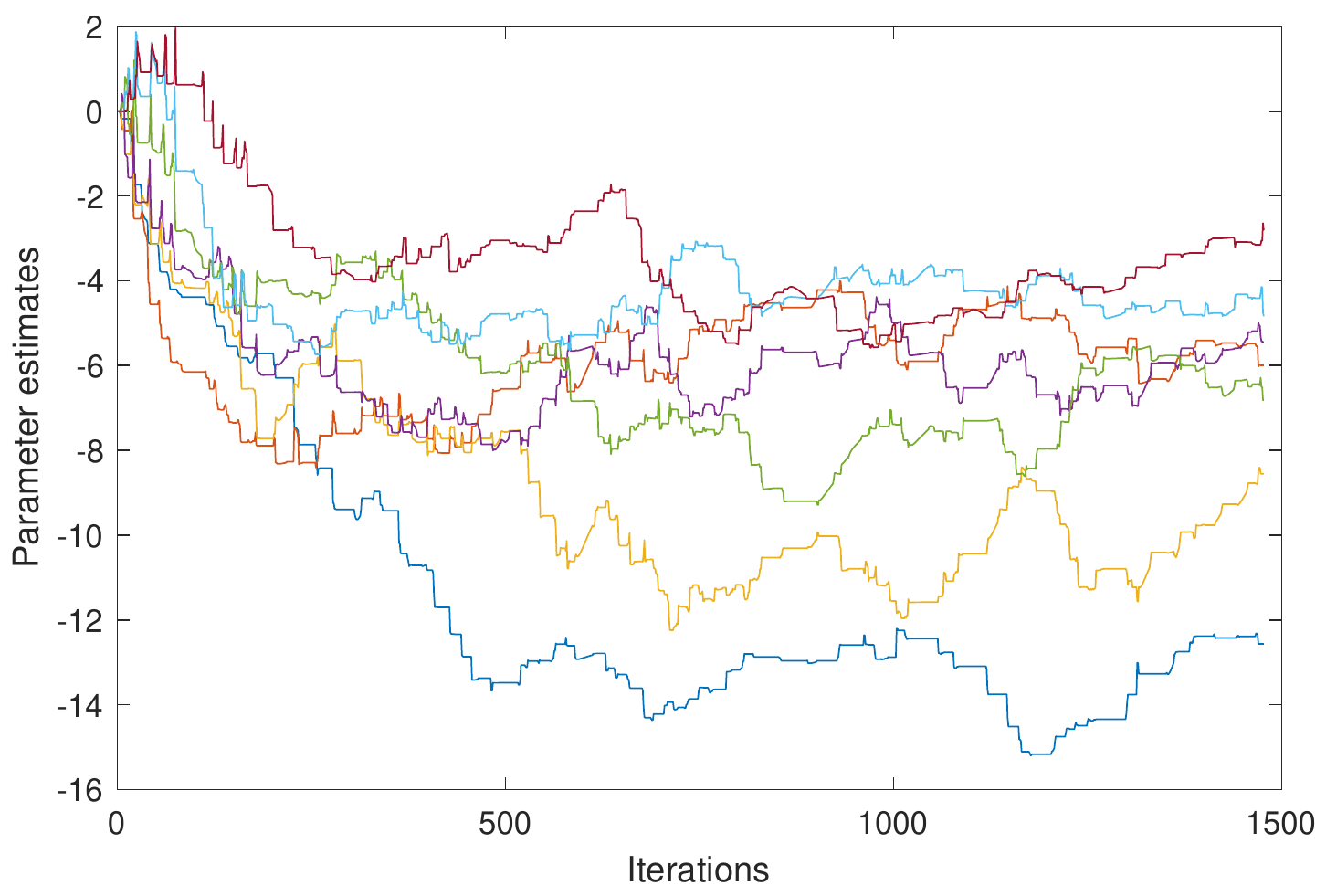}
\end{center}
\caption{Parameter estimates obtained in single-agent case using GTD2($\lambda$) algorithm. Each color corresponds to a different parameter (with 7 parameters total).}
\label{single_agent}
\end{figure}

In the second experiment we demonstrate the denoising effect of the introduced distributed algorithms. We assume that the agents have the same stationary behavior policies as above, but that they all start in state 1 and are able to advance to the final state 15. We also assume that the agents locally implement the algorithms with eligibility traces, with different $\lambda$ parameters: $[0.6, 0.1, 0.25, 0.5, 0.05, 0.01, 0.3, 0.5, 0.4, 0.7]$. In Fig.~\ref{dgtd2_value}, the value function approximation obtained using D1-GTD2($\lambda$) for $\alpha=\beta=0.3$ (one time scale) is represented. It can be seen that the approximation is better for the states $z_i\in \{1,3,5,7,9,11,13\}$, since these are references for the radial basis representation (note that it is not possible to converge to the true value function because of the introduced function approximation). As can be seen from the figure, all the agents have achieved consensus: the final value function approximations are practically the same for all the agents. Fig.~\ref{dgtd2_parameters} shows the parameter estimates $\theta_i(n)$ as functions of the number of iterations $n$. Note that, in this case, 20 episodes were needed for the obtained approximation, which is much less compared to the single agent case (Fig.~\ref{single_agent}), which also has much larger variance. Note that, in the case of fully connected network (centralized case), the improvements (rate of convergence, agents' agreement, denoising) are very slight compared to the case of sparsely connected network (as expected based on Fig.~\ref{fig_net_comp1}).

\begin{figure}
\begin{center}
\includegraphics[width=0.6\columnwidth]{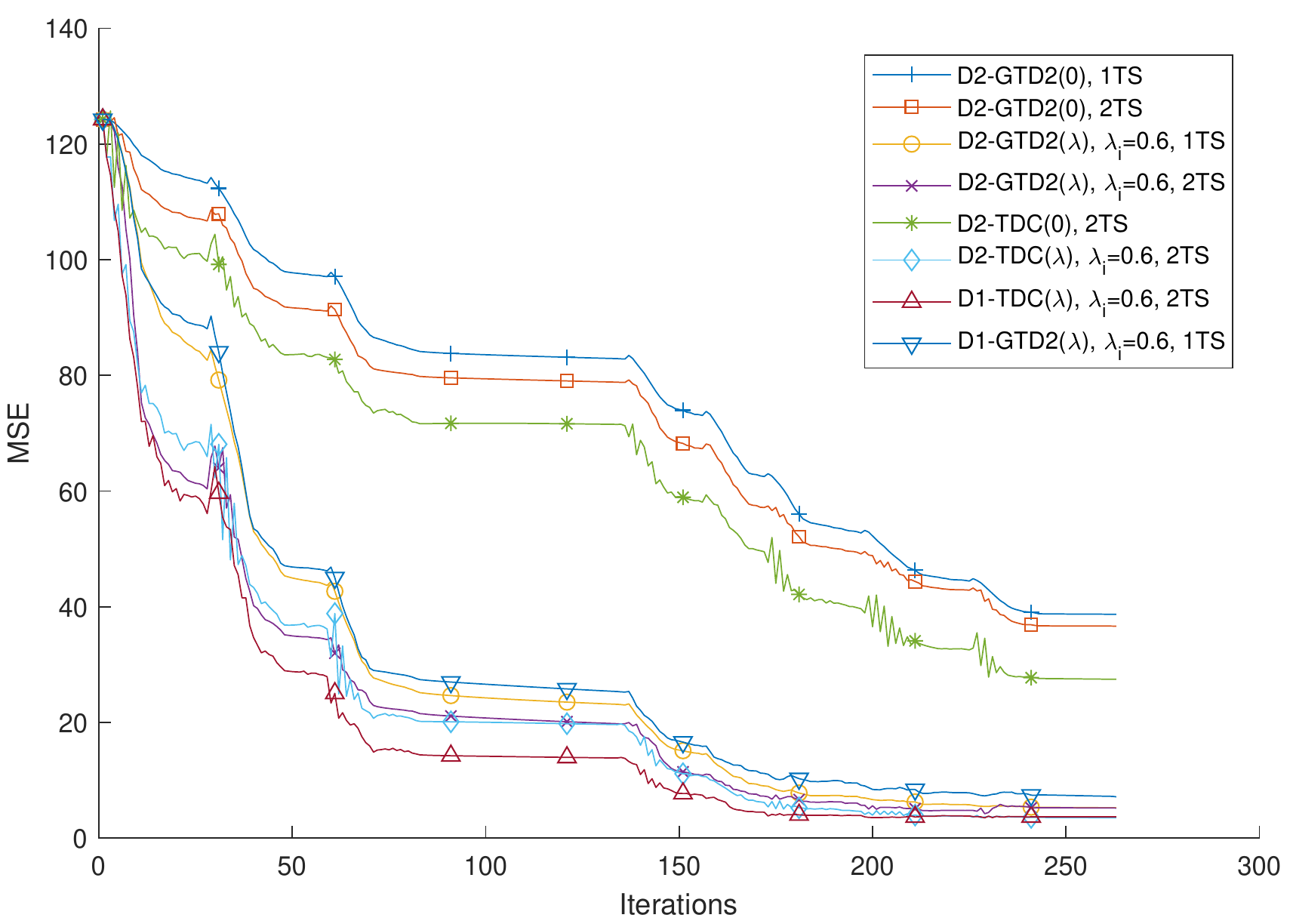}
\end{center}
\caption{Comparison of the mean-square error of value function approximation obtained by using the proposed schemes as well as the baseline from \cite{valcarcel2015distributed,naira_cdc_2018,doan_finite_time_2019} (corresponding to the D2-GTD(0) in one time scale). The baseline has been outperformed by our algorithms, with D1-TDC($\lambda$) having the best convergence rate.}
\label{fig_comparison}
\end{figure}

Finally, we have performed a test comparing the performance of all the proposed algorithms. Note that the algorithms previously proposed and analyzed in \cite{valcarcel2015distributed,naira_cdc_2018,doan_finite_time_2019}, which can serve as a baseline, are actually a special case of our work, corresponding to our D2-GTD($\lambda$) for $\lambda=0$ (no eligibility traces) and implemented in one time scale. For the same setup as in the previous experiment, we run the following 8 algorithms: 1) D2-GTD(0), one time scale (TS), 2) D2-GTD(0), two TS, 3) D2-GTD($\lambda$), $\lambda_i=0.6$, one TS, 4) D2-GTD($\lambda$), $\lambda_i=0.6$, two TS, 5) D2-TDC(0), two TS, 6) D2-TDC($\lambda$), $\lambda_i=0.6$, two TS, 7) D1-TDC($\lambda$), $\lambda_i=0.6$, two TS, and 8) D1-GTD($\lambda$), $\lambda_i=0.6$, $i=1,...,10$, one TS. We assume zero initial conditions for all the parameters, and compare the rate of convergence of the value function approximations. Fig.~\ref{fig_comparison} shows the mean-square error of the obtained value function approximations with respect to the true one. It can be seen that the baseline algorithm, proposed in \cite{valcarcel2015distributed,naira_cdc_2018,doan_finite_time_2019}, has the worst performance, while our D1-TDC($\lambda$) (for $\lambda_i=0.6$, with two time scales) has the best performance. Furthermore, in general, the algorithms with eligibility traces ($\lambda_i=0.6$) have better performance than with $\lambda_i=0$, and two-time-scale versions also increase the rate of convergence (note that TDC algorithm only works in two time scales). What can also be observed is that, at least in this problem setup, the algorithms without consensus on $w$ (D1-types of the proposed algorithms) have slight advantage over the D2-type algorithms.

\section{Conclusion}

In this paper we have proposed several novel algorithms for distributed off-policy gradient based value function approximation in a collaborative multi-agent reinforcement learning setting characterized by strict ISC. The algorithms are based on integration of stochastic time-varying dynamic consensus schemes into local recursions based on off-policy gradient temporal difference learning, including state-dependent eligibility traces. The proposed distributed algorithms differ by the algorithm form, by the choice of time scales and by the way the consensus iterations are incorporated. Under nonrestrictive assumptions, we have proved, after formulating asymptotic mean ODE's for the algorithms, that the parameter estimates weakly converge to consensus. The proofs themselves represent the major contribution of the paper. Contributions encompass an analysis of the asymptotic convergence rate and a demonstration of ``denoising" resulting from consensus. Furthermore, we have presented a discussion on the design of the communication network ensuring appropriate convergence points. Possibilities of direct extension of the results to the case of weak ISC have been indicated. Finally, efficiency of the proposed algorithms have been illustrated by numerous simulations.

Further work could be devoted to the weak convergence analysis of alternative multi-agent temporal difference schemes, including the emphatic temporal-difference algorithm \cite{emph,yu_2016} and actor-critic algorithms \cite{zhang2018fully,basar2019}. Also, the proposed schemes could be extended to the cases of nonlinear value function approximations (such as using deep neural networks \cite{OroojlooyJadid_review_2019}).

\section*{Acknowledgments}
This research was supported by the Science Fund of the Republic of Serbia, Grant \#6524745, AI-DECIDE, and by the Funda\c{c}\~{a}o para a Ci\^{e}ncia e a Tecnologia under Project UIDB/04111/2020.

\bibliography{marl3}

\end{document}